\documentclass[10pt,twocolumn,letterpaper]{article}

\usepackage{cvpr}              

\usepackage[utf8]{inputenc} 
\usepackage[T1]{fontenc}    
\usepackage{url}            
\usepackage{booktabs}       
\usepackage{amsfonts}       
\usepackage{nicefrac}       
\usepackage{microtype}      
\usepackage{xcolor}         

\usepackage{amsmath,bbm,bm,graphicx,amsthm}
\usepackage{subcaption}

\usepackage{algorithm}
\usepackage{algorithmic}
\usepackage{multirow}
\usepackage{pifont}

\definecolor{cvprblue}{rgb}{0.21,0.49,0.74}
\usepackage[pagebackref,breaklinks,colorlinks,allcolors=cvprblue]{hyperref}

\def\paperID{*****} 
\def\confName{CVPR}
\def\confYear{2025}

\title{A Simple and Efficient Baseline for Zero-Shot Generative Classification}

\author{Zipeng Qi$^{1,\ast}$,
Buhua Liu$^{1,\ast}$,
Shiyan Zhang$^{2}$, Bao Li$^3$, Zhiqiang Xu$^4$, Haoyi Xiong$^5$ and Zeke Xie$^{1,\dagger}$ \\
\vspace{3pt}\\
$^1$xLeaF Lab, The Hong Kong University of Science and Technology (Guangzhou),\\
$^2$Tsinghua University,
$^3$Institute of Automation, Chinese Academy of Sciences,\\
$^4$Mohamed bin Zayed University of Artificial Intelligence,
$^5$Baidu\\
}

\begin{document}
\maketitle
\renewcommand{\thefootnote}{\fnsymbol{footnote}} 
\footnotetext[1]{These authors contributed equally to this work.} 
\footnotetext[2]{Correspondence to: \textit{zekexie@hkust-gz.edu.cn}} 

\begin{abstract}
Large diffusion models have become mainstream generative models in both academic studies and industrial AIGC applications. Recently, a number of works further explored how to employ the power of large diffusion models as zero-shot classifiers. While recent zero-shot diffusion-based classifiers have made performance advancement on benchmark datasets, they still suffered badly from extremely slow classification speed (e.g., $\sim$1000 seconds per classifying single image on ImageNet). The extremely slow classification speed strongly prohibits existing zero-shot diffusion-based classifiers from practical applications. In this paper, we propose an embarrassingly simple and efficient zero-shot Gaussian Diffusion Classifiers (GDC) via pretrained text-to-image diffusion models and DINOv2. The proposed GDC can not only significantly surpass previous zero-shot diffusion-based classifiers by over 10 points ($61.40\% \to 71.44\%$) on ImageNet, but also accelerate more than 30000 times ($1000 \to 0.03$ seconds) classifying a single image on ImageNet. Additionally, it provides probability interpretation of the results. Our extensive experiments further demonstrate that GDC can achieve highly competitive zero-shot classification performance over various datasets and can promisingly self-improve with stronger diffusion models. To the best of our knowledge, the proposed GDC is the first zero-shot diffusion-based classifier that exhibits both competitive accuracy and practical efficiency. 

\end{abstract}

\section{Introduction}
\label{sec:intro}

Diffusion models, as the representative of generative models, have achieved tremendous success and attracted great attention due to their impressive performance \citep{dhariwal2021diffusion}. Diffusion models have been proposed with similar ideas, including Diffusion Probabilistic Models \citep{sohl2015deep}, Noise-conditioned Score-based Generative Models \citep{song2019generative}, and Denoising Diffusion Probabilistic Models \citep{ho2020denoising}. Diffusion models serve as backbone algorithms of various AIGC applications, including image generation \citep{nichol2021glide,rombach2022high,saharia2022photorealistic, qi2024not, qi2023layered}, video generation \citep{ho2022imagen,blattmann2023stable}, 3D generation \citep{poole2022dreamfusion,lin2023magic3d,xu2023dream3d}, and speech generation \citep{kong2020diffwave,yang2023diffsound, qi2023difftalker}.

Diffusion models have been continually improving with increasingly more model parameters, training data, and computational resources \citep{peebles2023scalable,podell2023sdxl}. We particularly note that modern large generative models usually have cost much more computational resources and training data than modern discriminative models. With such strong generative ability and large resources/data, large generative models should be very powerful foundation models for more applications. A generative classifier is a type of classification model that is based on the principles of generative models. This line of research, often called \textit{zero-shot generative classifiers}, has suggested promising potential and advantages, especially in the era of large generative models. Recent advancements in large diffusion models \citep{podell2023sdxl,sauer2023adversarial,yang2023diffusion, qi2023layered} make powerful zero-shot generative classifiers become more possible.

Only a few pioneering works have explored \textit{zero-shot diffusion-based classifiers}. \citet{li2023diffusion} and \citet{clark2023text} similarly employed the denoising loss of each class via large Stable Diffusion (SD) models to achieve state-of-the-art zero-shot classification performance on ImageNet \citep{deng2009ImageNet}. \citet{jaini2023intriguing} studied intriguing properties of zero-shot diffusion-based classifiers and argued that they approximate human object recognition data surprisingly well. Emerging zero-shot diffusion-based classifiers are mainly loss-based diffusion-based classifiers. While we have seen progress in recent works, zero-shot diffusion-based classifiers are still very naive and largely suffer from extremely slow classification speed. For example, it takes around 1133 seconds per classifying a single image from ImageNet \citep{li2023diffusion}. The poor efficiency is exactly the main bottleneck of applying zero-shot diffusion-based classifiers widely in practice. 

\textbf{Contribution} In this paper, we solve the efficiency problem of zero-shot diffusion-based classifiers and demonstrate their promising performance. We made two contributions. 

\begin{figure*}[t]
\centering
\includegraphics[width=1.0\textwidth]{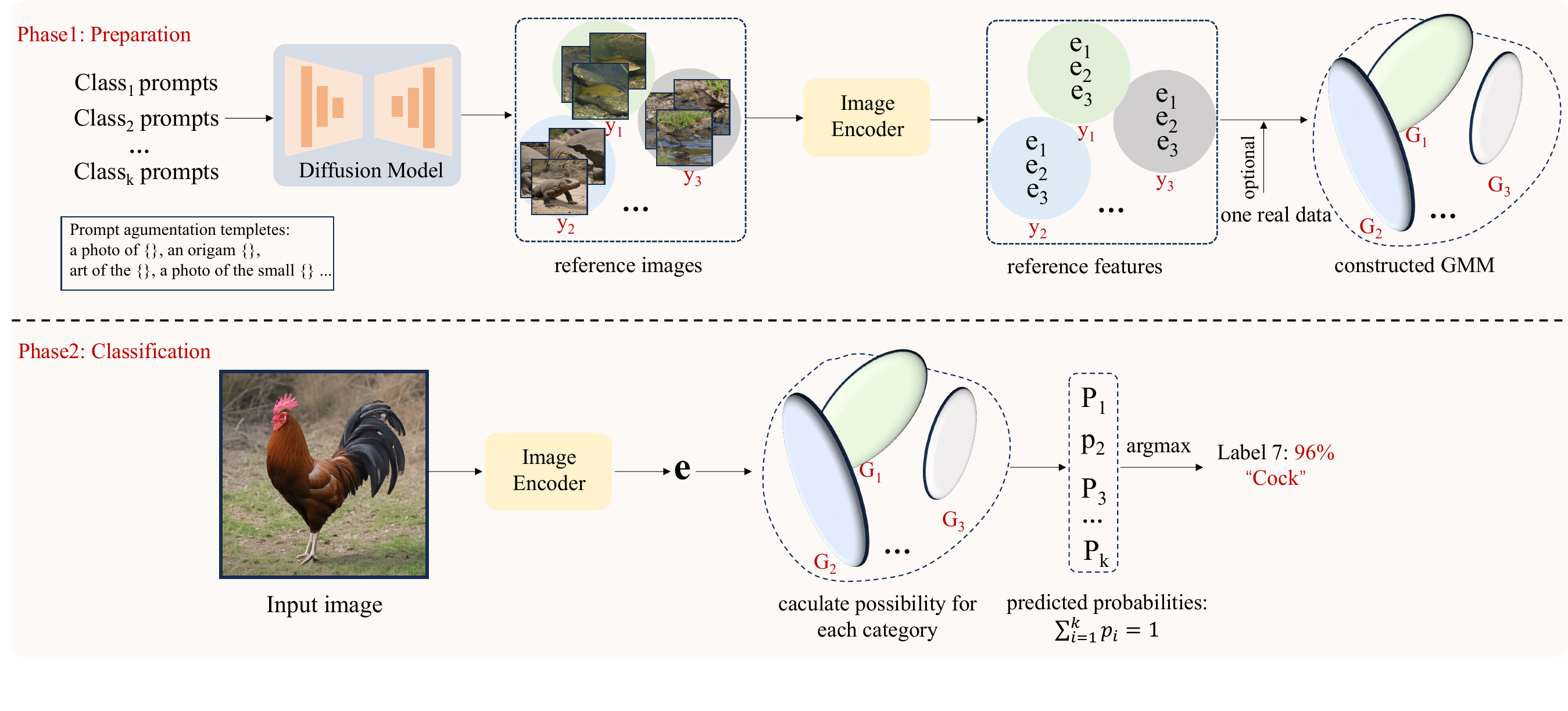}  
\caption{The overview of our Gaussian Diffusion Classifiers (GDC), which consists of two phases: 1) Preparation Phase and 2) Gaussian-based Classification Phase. GDC not only can output the classification result but also the probability. For conciseness, we present only three reference images, while the influence of varying numbers of reference images is discussed in Section~\ref{sec:empirical}. Algorithm~\ref{algo:gdc} shows more details.}
\label{fig:overview} 
\end{figure*}

First, we introduce a simple yet useful baseline Diffusion-based zero-shot classifier, namely the Gaussian Diffusion Classifier (GDC), which can predict the class probability give a test image efficiently and accurately. We present its illustration in Figure \ref{fig:overview} and its pseudocode in Algorithm \ref{algo:gdc}. It accelerates more than 30000 times (from $1133$ seconds to $0.03$ seconds), classifying a single image on ImageNet during evaluation. To the best of our knowledge, the proposed GDC is the first zero-shot diffusion classifier that exhibits both competitive accuracy and practical efficiency. 

Second, our extensive experiments demonstrate that GDC can achieve impressive zero-shot classification performance over various datasets. For example, as Table \ref{table:gdcbenchmark} shows, the proposed GDC can significantly surpass previous zero-shot diffusion-based classifiers by more than 10 points (from 61.40\% to 71.44\%) on ImageNet, the most popular image classification benchmark. 
Interestingly, when provided with one single training images, GDC can outperform CLIP.
We also show that GDC can promisingly self-improve its performance with stronger diffusion models, which will naturally appear in the future.

\textbf{Structure of the paper} In Section \ref{sec:related}, we review  related work and highlights distinctions from zero-shot diffusion-based classifiers. In Section~\ref{sec:motivation}, we clearly explain our motivation for using Gaussian-based classifier. In Section \ref{sec:method}, we formally introduce our GDC method details. In Section \ref{sec:empirical}, we present comprehensive experimental results, demonstrating both the empirical success and unique properties of our zero-shot diffusion-based classifiers. In Section \ref{sec:discuss} we provide case analyses and discusses the main limitations, while in Section \ref{sec:conclusion} we conclude our work.

\section{Related Work}
\label{sec:related}

In this section, we review related work and discuss how they differ from the studies of zero-shot diffusion-based classifiers.

\textbf{Text-to-Image Diffusion Models} Since generative diffusion models have been proposed, pioneering works have shown impressive image generation improvements over previous mainstream GAN-based methods \citep{dhariwal2021diffusion,ramesh2021zero}. GLIDE \citep{nichol2022glide} is the pioneering work on text-to-image generation, which explored classifier-free guidance for the sample photorealism and caption similarity. Imagen \citep{saharia2022photorealistic} further adopted a pretrained and frozen large language model as the text encoder. Latent Diffusion is a milestone of text-to-image diffusion models. SD \citep{rombach2022high} made significant progress towards diffusion in low-dimensional latent space. Dall-E 2 \citep{ramesh2022hierarchical} further explored multimodal latent space. We recommend \citet{yang2023diffusion} for a comprehensive survey on diffusion models. However, this line of research only focused on generative tasks and failed to explore the promising potential of diffusion models in discriminative tasks.

\textbf{Generative classifiers} Discriminative models directly learn to model the decision boundary of the underlying classification task, while generative models learn to model the (conditioned) distribution of the data. A large body of works \citep{ng2001discriminative,li2019generative,sensoy2020uncertainty,van2021class,zheng2023revisiting} studied generative classifiers from a Bayesian perspective. A number of works \citep{lee2019robust,croce2020gan,mackowiak2021generative,zimmermann2021score} reported that generative classifiers may naturally accomplish relatively good robustness for classification. Chen~\citep{chen2023robust} highlights the generative classifiers with pre-trained diffusion models have potential for adversarial robustness compared with the commonly studied discriminative classifiers. Several seminal works \citep{bucher2017generating,hjelm2018learning,brown2020language,he2022masked,burgert2022peekaboo} studied efficient representations of generative models, which are later used for downstream tasks. However, directly employing generative models for classification in a zero-shot way is still under-explored. Most of these works required training data and did not touch zero-shot generative classification. Zero-shot generative classifiers require essentially different methods and exhibit novel intriguing properties \citep{jaini2023intriguing} beyond conventional generative classifiers that require training data.

\textbf{Zero-Shot Classification}  Common machine-learning methods focus on classifying instances whose classes have already been seen in training, which is sometimes not possible in practice. Zero-shot learning is a powerful and promising learning paradigm where, at test time, a learner observes samples from classes that were not observed during training and needs to predict the class that they belong to \citep{wang2019survey}. Some works have explored zero-shot classification with various techniques, such as co-occurrences of visual concepts for knowledge transfer \citep{mensink2014costa}, learning latent embeddings \citep{xian2016latent,jiang2017learning,ye2017zero}, and generating visual representations \citep{bucher2017generating}. Recently, it has become well known that large vision-language models, such as CLIP, are robust and strong unsupervised multitask learners \citep{radford2019language,wortsman2022robust}. Further, fine-tuned large language models are also reported to be powerful zero-shot learners for many downstream tasks \citep{wei2021finetuned}. 

This naturally leads to an open question: are large diffusion models also efficient and  powerful zero-shot classifiers? Moreover, if large diffusion models are capable, how to enable the model output the classification results and further endow the results with a probabilistic interpretation? Large diffusion models must require essentially different zero-shot methodology beyond the previous zero-shot methodology of Large Language Models. As we mentioned in Section \ref{sec:intro}, this line of research only contains a few pioneering works and is still largely under-explored.

\begin{algorithm}[th]
 \caption{Gaussian Diffusion Classifier}
 \label{algo:gdc}
\begin{algorithmic}
   \STATE {\bfseries Input:} examples to classify $\{x_{j}\}_{j=1}^{m}$, data categories $\{y_{i}\}_{i=1}^{k}$, a diffusion model $\mathcal{M}$, an image encoder $\mathcal{E}$,  $N$ reference images per class
   \STATE \textcolor{purple}{1) Preparation Phase} 
   \STATE {\bfseries Initialize} $\mathrm{embeddings}[y_{i}] = []$ for each class 
   \FOR{$i=1$ {\bfseries to} $k$} 
   \FOR{$k=1$ {\bfseries to} $n$} 
   \STATE $\tilde{x} = \mathcal{M}(y)$ // Generate reference images
   \STATE $e = \mathcal{E}(\tilde{x})$ // Compute image embeddings 
   \STATE $\mathrm{embeddings}[y_{i}]$.append$(e)$ // Store image embeddings for each class/cluster
   \ENDFOR
   \ENDFOR
   \STATE GMM = $[G_1, G_2 \dots G_k]$// Construct GMM with $k$ Gaussian components using stored clusters
   \STATE \textcolor{purple}{2) Gaussian-based Classification Phase}
   \FOR{$j=1$ {\bfseries to} $n$}
   \STATE $e = \mathcal{E}(x_{j})$ // Compute test images' embeddings 
   \STATE {\bfseries Initialize} $\mathrm{probility}[y_{i}] = 0$ for each cluster
   \FOR{$i=1$ {\bfseries to} $k$}
   \STATE // Decide which cluster $x_{j}$ belongs to
   \STATE $p(y_i|x) = \frac{p(e|y_i)p(y_i)}{\sum_{j=1}^kp(e|y_j)p(y_i)}$
  
   \ENDFOR
   \STATE  $x\in y_i, \iff y_i = \mathop{\arg\max}\limits_{y_i} p(e|y_i)$ // Predict the label by the maximal probability
   \ENDFOR
\end{algorithmic}
\end{algorithm}

\section{Motivation}

\label{sec:motivation}
\begin{figure}
    \centering
    \includegraphics[width=1.0\linewidth]{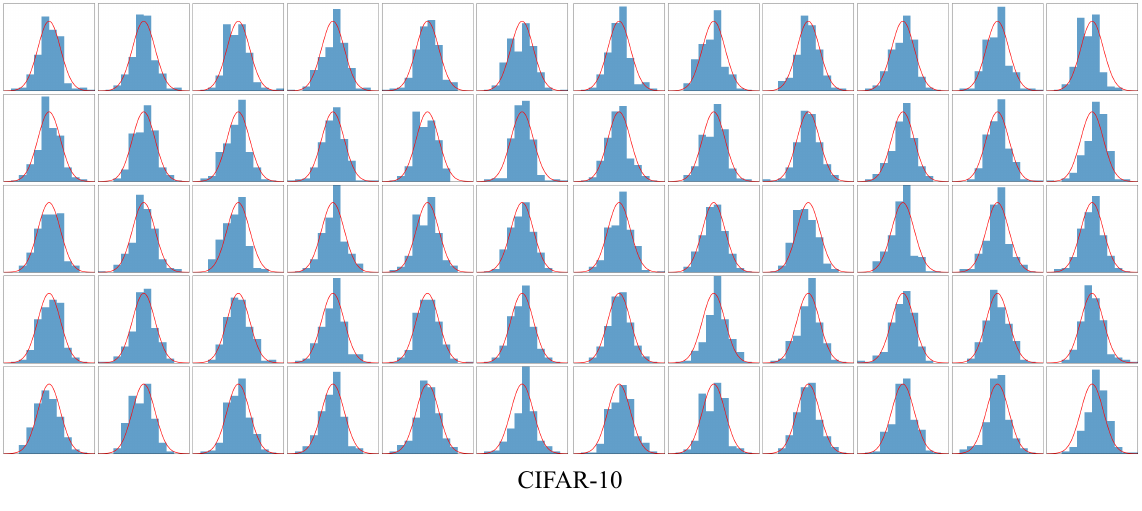}
    \caption{We visualize the distribution of image features for random 60 PCA components from a randomly picked class on CIFAR-10. The red curves representing fitted Gaussian distributions. The results indicate that the feature values approximate a Gaussian distribution, motivating our method.}
    \label{fig:motivation}
    \vspace{-0.3cm}
\end{figure}
We use DINOv2~\citep{oquab2023dinov2} to extract image features from two well-known classification datasets, CIFAR100 and ImageNet, and analyze the resulting feature distributions. We examine the distribution of random 30 components from PCA. As shown in Figure~\ref{fig:motivation}, the feature values approximate a Gaussian distribution. We assess the Gaussianity of image features of each class in ImageNet by the Shapiro-Wilk normality test \citep{shapiro1965analysis}.
The statistical test results show that, for the 1,000 classes in
ImageNet, more than 60\% principal components’ p-value are
higher than 0.05, which indicate that these components can
be approximately Gaussian. Moreover, some components’
p-values are even higher than 0.99. Appendix C and D show more
visual and statistical results. The visualization and statics results motivate our natural use of a Gaussian Mixture Model (GMM) to represent each class’s feature distribution. In this framework, each Gaussian component captures the mean and variability of the embeddings for a specific class and predict the modeling probability of a given image embedding. By modeling class distributions probabilistically, the GMM not only improves classification accuracy but also 194 enables a probabilistic interpretation, facilitating uncertainty quantification and enhanced interpretability of the results.


\section{Methodology}
\label{sec:method}

In this section, we present our Gaussian Diffusion Classifiers (GDC) and show why it can solve the efficiency problem of zero-shot diffusion-based classifiers. 

\textbf{Notations} Suppose we try to classify samples from a dataset $\mathcal{D}$ which contain $k$ categories, $\{y_{i}\}_{i=1}^{k}$, and $m$ samples, $\{x_{j}\}_{j=1}^{m}$. A diffusion model $\mathcal{M}$, which models the text-conditioned distribution $p(x|c)$, can generate images conditioned on a given class caption $c$.

Previous zero-shot diffusion-based classifiers \citep{li2023diffusion,clark2023text} are loss-based methods. When performing classification, they must spend expensive costs doing diffusion reverse inference thousands of times per classifying a single image to estimate some diffusion-based loss of the image $x$ and each class. If a diffusion model produces a relatively low loss of the pair of the image $x$ and the class caption $c$, it suggests that the class caption $y$ may describe the image $x$ better than the other candidate class caption. This strategy is natural but extremely slow, especially when the number of classes is large. For example, this strategy must repeat $k=100$ times and $k=1000$ times per classifying a single image from CIFAR100 and ImageNet, respectively.

To solve the efficiency problem of zero-shot diffusion-based classifiers, we must avoid repeating diffusion backward processes $k$ times per classifying a single image. The basic idea of our method is very simple but effective. We present the pseudocode of GDC in Algorithm \ref{algo:gdc}. Our GDC method consists of two phases. 

\textbf{1) Preparation Phase}. This phase includes reference image generation and GMM construction. Before classifying any images, we use diffusion models to generate $n$ (e.g., 240) reference images per class. To increase the diversity of these reference images, we apply a prompt augmentation strategy~\citep{osowiechi2024watt}. This approach allows us to generate only $N \times k$ images to classify arbitrarily large image sets, including over one million images from ImageNet. We then use DINOv2~\citep{oquab2023dinov2} to extract $d$-dimensional embeddings for all reference images. This results in $k$ clusters, each containing $N$ data points in the embedding space, which we use to construct a Gaussian Mixture Model (GMM) with $k$ components. 
For each Gaussian component, we calculate the cluster mean vector $\mu$ and the cluster covariance matrices $\Sigma$. For the GMM, we combine the mean vectors as $[\mu_1, \mu_2, \dots \mu_k] \in \mathcal{R}^{k\times d}$ and covariance matrices as $[\Sigma_1, \Sigma_2, \dots, \Sigma_k] \in \mathcal{R}^{k\times d \times d}$. 
We calculate the cholesky decomposition of the precision matrixes, the inverse of covariance matrix for the GMM. 
\begin{equation} 
\mathrm{cholesky}([\hat{\Sigma}_0, \hat{\Sigma}_1,\dots \hat{\Sigma}_k]^{-1}) = \mathbf{LL}^*, 
\end{equation}
where $\mathbf{L}$ is a lower triangular matrix with real and positive diagonal entries, and $\mathbf{L}^*$ denotes the conjugate transpose of $\mathbf{L}$. Appendix C presents the calculate method for $\mathbf{L}$. The matrix \( \hat{\Sigma}_i \) is defined as \( \Sigma_i + \epsilon I \), and its inverse, \( \hat{\Sigma}_i^{-1} \), is referred to as the precision matrix. Here, $I$ represents identity matrices with the same shape as $\Sigma$. The regular term, $\epsilon$ , should be a small positive value (e.g. $10^{-8}$), under the premise of ensuring the positive certainty of the precision matrix. Appendix B presents the relevant experimental results for regularization value selection. It is note that storing the precision matrices instead of the covariance matrices makes it more efficient to compute the log-likelihood of new samples at test time. 


\textbf{2) Gaussian-based Classification Phase}. 
Given a set of cluster data includes $k$ classes, the task of a classifier is to assign a sample $x$ to one of the k classes. 
According to Bayes rule, $p(y_i|x)$ can be calculated by using the class-conditional probability $p(x|y_i)$ and the class prior probability of the class $p(y_i)$: $p(y_i|x) = \frac{p(x|y_i)p(y_i)}{p(x)}$. Thus, the main idea of a classifier based on Bayes decision theory can be written as flowing:
\begin{equation}
    x\in y_i, \iff y_i = \mathop{\arg\max}\limits_{y_i}p(x|y_i)p(y_i).
\end{equation}
The class prior probability $p(y_i)$ is commonly known in advance or can be calculated/estimated using some approaches~\citep{lawrence2002neural, du2014class}. Thus, when we do zero-shot classification on an image $x$, we only need to use GMM to compute the class-condition probability $p(x|y_i)$ of $x$ and then assign $x$ to the class with the highest probability. Noticed, we firstly use the DINO to extract the features $e$ of $x$ and the input it to each Gaussian Model to calculate the probabilities. Thus, ther are $p(x|y_i) \iff p(e|y_i)$ and $p(y_i | x) \iff p(y_i|e)$.

Our GDC can give the probability interpretation for the results. The posterior probability can be calculated based on Bayes rule as flowing:
\begin{equation}
\begin{aligned}
p(y_i | x) \iff p(y_i|e) &= \frac{p(e | y_i) \, p(y_i)}{p(e)} \\
           &= \frac{p(e | y_i) \, p(y_i)}{\sum_{j=1}^k p(e |y_j) \, p(y_j)}.
\end{aligned}
\end{equation}
The $p(e|y_i) = \frac{1}{2\pi^{d/2}(|\Sigma_i|)^{1/2}} \exp\left(-\frac{1}{2}(e-u_i)^T\hat{\Sigma}_i^{-1}(e-u_i)\right)$ is the Gaussian density of the $i_{th}$ component in $d$-dimensional space, where $\hat{\Sigma}_i^{-1}$ is the $i_{th}$ precisions\_matrix. So $\sum_{i=1}^k p(y_i|x) = 1$. This can enable our GDC further analysis in uncertainty quantification, robustness evaluation, and risk assessment and also enhance the interpretability of the method.

The proposed GDC can classify an image very quickly during deployment, because the Gaussian-based Classification Phase only requires the forward process of the image encoder once and probability calculation. The expensive and slow diffusion reverse process only happens in the first Preparation Phase that generates clusters, which can be completed before deploying GDC. The first phase is prepared for the whole classification task rather than a single image. So our classification speed can be several orders of magnitudes faster than the conventional redundant diffusion reverse processes per classifying one image. 

\begin{table*}[th!]
\begin{center}
\begin{small}
\resizebox{0.9\textwidth}{!}{%
\begin{tabular}{c| ccccccccc | c}
\toprule
   Method& ImageNet  & CIFAR-10 & CIFAR-100 &  STL-10 & Pet-37 & Food-101  & Flower-102 & Caltech-101 &  DTD & Mean \\
\midrule
Li's DC \citep{li2023diffusion} &  61.4 & 88.5 & 67.6  &  95.4 & 87.3 & 77.7 & 66.3  & - &  - & 77.7 \\
Clark's DC \citep{clark2023text} &  61.9  & 72.1 & 45.3 &  92.8  & 72.5  & 71.6 & - & 73.0 & 44.6 & 66.7 \\
GDC (Ours) & \textbf{71.4} & \textbf{96.8} & \textbf{84.0} & \textbf{96.5} & \textbf{92.3} & \textbf{81.2} & \textbf{81.4} & \textbf{91.4} & \textbf{48.2}& \textbf{82.6} \\
\bottomrule
\end{tabular}
}
\end{small}
\end{center}
\caption{Zero-shot generative classification on benchmark datasets. The proposed GDC significantly outperforms the baselines.(we use the existing results to calculate the mean value.)}
\label{table:gdcbenchmark}
\end{table*}

\begin{table*}[th!]
\begin{center}
\resizebox{0.9\textwidth}{!}{
\begin{tabular}{c|c|ccc|c}
\toprule
 Dataset & Method & Preparation  & Dataset Classification & Total Time & Single-Image Classification \\
 \midrule 
\multirow{2}{*}{ImageNet}  & 
  GDC(Ours)  & 28.219 h & 0.410 h & 28.629 h & 0.030 s \\ 
& Li's DC \citep{li2023diffusion}  & 0 & 15737 h &  15737 h &  1133 s \\ 
\midrule
\multirow{2}{*}{CIFAR-100}  & GDC(Ours)  & 0.177 h & 0.052 h & 0.229 h & 0.019 s  \\ 
 & Li's DC \citep{li2023diffusion}& 0  & 333 h & 333 h & 120 s  \\ 
\bottomrule
\end{tabular}
}
\end{center}
\caption{Computational Time Comparison (V100 GPU time). Datasets: ImageNet and CIFAR-100. Single-image classification time reflects the efficiency of employing diffusion-based classifiers in the real world, while the total time reflects the efficiency of performing classification on the benchmark datasets.}
\label{table:time}
\end{table*}

\section{Empirical Analysis}
\label{sec:empirical}
In this section, we conduct extensive experiments and demonstrate the impressive efficiency and performance of GDC over the baseline methods.

\textbf{Experimental Settings} We use popular real-world classification datasets as benchmark datasets and open-source SD models as the backbone diffusion model of GDC. We hope to fairly compare our GDC with the previous state-of-the-art baseline, namely loss-based diffusion-based classifiers \citep{li2023diffusion,clark2023text}. We use SDXL-turbo as the default diffusion model to generate 240 reference images with $512 \times 512$ and DINOv2~\citep{oquab2023dinov2} as the default image encoder to output 1536-d vector for each reference image, unless we specify them otherwise. In this way, people can conveniently employ our GDC method in practice and reproduce our results for further research. The reference images are resized from $512\times 512$ to $224\times 224$ to align with the input dimensions of DINOv2. The default regularization value is $1e^{-8}$.

\textbf{Dataset} ImageNet \citep{deng2009ImageNet}, CIFAR-10/100 \citep{krizhevsky2009learning}, Flower-102 \citep{nilsback2008automated}, Oxford Pet-37 \citep{parkhi2012cats}, Food-101 \citep{bossard2014food}, STL10~\citep{coates2011analysis}, DTD~\citep{cimpoi14describing} and Caltech101~\citep{fei2006one}. More experimental details can be found in Appendix A.1.

\textbf{Prompt Argumentation} In the default setting, we use SDXL-turbo to generate $N=240$ reference images for each class, employing eight different prompt templates to increase the diversity of the reference images, such as `a photo of a \{\}', `an origami \{\}', `art of the \{\}', `a photo of the small \{\}', and so on(see Appendix A.2). We replace the placeholder {} with the name of the category. We use each prompt template to generate 30 reference images averagely.

\textbf{Zero-shot Generative Classification} We empirically compare the proposed GDC with the previous state-of-the-art baseline, Li's Diffusion Classifier(DC)~\citep{li2023diffusion} and Clark's DC~\citep{clark2023text} over various popular benchmark datasets. 

We note that, due to the extremely poor efficiency, the original works that proposed diffusion classifiers \citep{li2023diffusion,clark2023text} only used a subset ($\sim30000$ images) of test sets for evaluation. We reproduce the compared DCs performance on CIFAR-100, which was not reported by \citet{li2023diffusion} and directly use other numerical results of SD models in original papers \citep{li2023diffusion,clark2023text}, as compared DCs may not classify these whole datasets within a reasonable time. As code relase problem, we may not present the numerical results of \citet{clark2023text} on Flower-102. Also, due to the extensive testing time required by Li's DC~\citep{li2023diffusion}, we are unable to test it on DTD and Caltech-101 within a reasonable time. Instead, we use the existing results to compute the mean values in Table~\ref{table:gdcbenchmark}.

The results in Table \ref{table:gdcbenchmark} demonstrate our GDC significantly outperforms the baselines on benchmark datasets. The improvements on the three most popular datasets, including ImageNet, CIFAR-10/100, and Flowers-102, are especially significant. We believe it is because these three datasets contain rich classes and real-world natural objects, which are good for generating rich and representative reference clusters. We report that GDC can consistently outperform compared DCs over all datasets. And, the mean accuracy gains of CGC are also up to \textbf{4.9} points and \textbf{15.9} points compared with Li's DC \citep{li2023diffusion} and Clark's DC \citep{clark2023text}, respectively, over the evaluated multiple datasets. Considering the efficiency advantage of GDC, the accuracy gains are very impressive.

\textbf{Computational Efficiency} Since poor efficiency is the main bottleneck of employing zero-shot diffusion-based classifiers, it is essentially important to analyze the computational efficiency. In Table \ref{table:time}, we carefully evaluate the computational efficiency of GDC and Li's DC on ImageNet and CIFAR-100. We particularly focus on single-image classification time, because this metric reflects the efficiency of employing diffusion-based classifiers in the real world. If the single-image classification time is short enough, zero-shot diffusion-based classifiers with good classification performance will truly start to compete with common discriminative models on classification tasks. We also present the total time of performing zero-shot classification on the benchmark datasets, because the time of classifying a dataset provides simple and fair efficiency comparisons. Fortunately, we find that the proposed GDC is several orders of magnitude faster than Li's DC in both efficiency metrics. 

\textbf{One-shot Classification Results} Table~\ref{table:gdcbenchmark} presents the zero-shot performance across various benchmarks, while Table~\ref{table:one-shot} demonstrates the accuracy on ImageNet, CIFAR10 and CIFAR100 under one-shot setting. We randomly replace one reference image with a real image. We employ pre-trained OpenCLIP(ViT-H-14~\citep{radford2021learning}) on these dataset to calculate the similarity score between the visual features and text features.The text prompt is 'a photo of a \{\}', with the \{\} replaced by the class name. The test image is assigned to the class with the highest similarity score. From the results, we observe that even with only one single real data, the performance of GDC surpasses CLIP over 1.0, 1.2 and 4.2 receptively on ImageNet, CIFAR10 and CIFAR100. Incorporating real data enables the estimated Gaussian model to more accurately capture the distribution of the test data.

\begin{table}[t]
\begin{center}
\begin{small}
\resizebox{0.48\textwidth}{!}{%
\begin{tabular}{c|ccc|c}
\toprule
 Method & ImageNet &CIFAR-10 &CIFAR-100   & Mean  \\
\midrule
CLIP & 75.2 & 97.2 & 84.2&  85.5 \\
GDC (zero-shot) & 71.4 & 96.8 & 84.0 & 77.7 \\
GDC (one-shot) & \textbf{76.2} & \textbf{98.4}& \textbf{88.4} & \textbf{87.7} \\

\bottomrule
\end{tabular}
}
\end{small}
\end{center}
\caption{One-shot generative classification accuracy of GDC. We randomly replace the generated reference images with one single training image.}
\label{table:one-shot}
\end{table}

\section{Ablation Analysis}
In this section, we analysis the key factors affecting GDC performance.

\begin{figure}
\begin{minipage}[t]{0.5\textwidth}
\includegraphics[width =1.\columnwidth ]{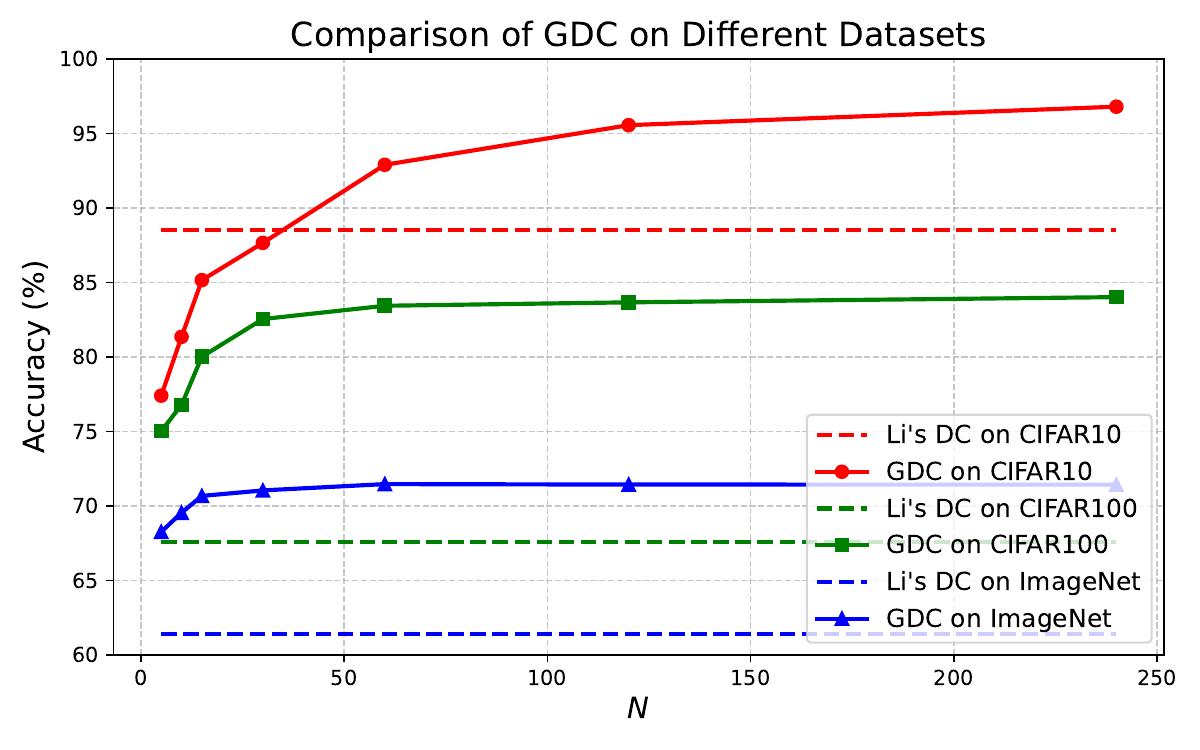} 
\caption{The test accuracy of GDC with various choices of $N$ on CIFAR10, CIFAR100 and ImageNet. The performance of GDC is pretty robust to the choice of $N$, when $N\geq 100$. Even if we have only one reference image per class, GDC performs still better than conventional Li's DC.}
\label{fig:gdc_k_curve}
\end{minipage} 
\hfill
\begin{minipage}[t]{0.5\textwidth}
\centering
\includegraphics[width =1.0\columnwidth ]{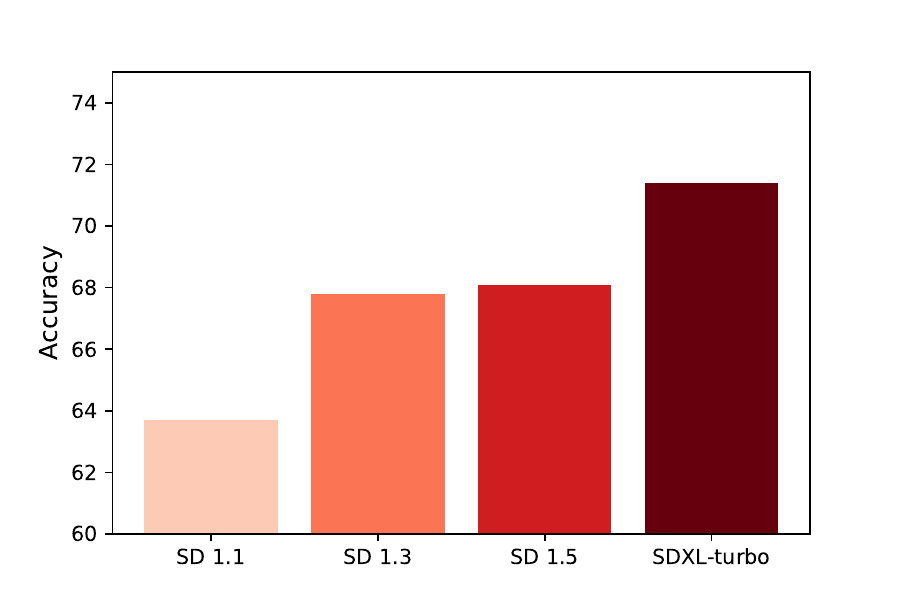}  
\caption{The classification accuracy of GDC monotonically increases with stronger diffusion models on ImageNet. Generation Performance: SDXL-turbo > SD 1.5 > SD 1.3 > SD 1.1. }
 \label{fig:gdc_stronger_diffusion}
 \end{minipage} 
\end{figure}

\textbf{Choices of the hyperparameter $N$}. $N$ as Algorithm \ref{algo:gdc} shows, we need to specify the choices of the hyperparameter $n$ (the number of reference images per class). We first conduct the experiment of GDC with $N \in \{5, 10, 30, 60,120,240\}$ in Figure \ref{fig:gdc_k_curve}. With more reference images, we can obtain more statistically stable clusters for zero-shot classification, which often leads to higher classification accuracy. While we use $N=240$ as the default setting, we notice that GDC with $100$ reference images can perform similarly well and cost only half preparation time while trying to maintain performance as much as possible. The results in the figure suggest that GDC is pretty robust to the choice of $N$. Of course, the choice of $N$ almost does not affect single-image classification time, because the embedding extraction of the image encoder costs most single-image classification time. 

\textbf{Without Prompt Augmentation} To evaluate the importance of the diversity of reference image, we compare the results of eights prompts and one single prompt, `a photo of a \{\}' on three representative datasets: ImageNet, CIFAR10 and CIFAR100.
\begin{table}[t]
\begin{center}
\begin{small}
\resizebox{0.48\textwidth}{!}{%
\begin{tabular}{c|ccc|c}
\toprule
 Method & ImageNet & CIFAR-10 & CIFAR-100  & Mean\\
\midrule
wo/ augmentation & 68.5 & 87.2 & 77.3 & 77.7 \\
w/ augmentation & \textbf{71.4} & \textbf{96.8} & \textbf{84.0} & \textbf{84.1} \\
\bottomrule
\end{tabular}
}
\end{small}
\end{center}
\caption{Zero-shot generative classification accuracy of GDC with vs. without the prompt augmentation strategy for reference images on ImageNet, CIFAR10 and CIFAR100 datasets.}
\label{table:diversity_prompt}
\end{table}

Table~\ref{table:diversity_prompt} clearly shows that the results with reference images generated using diversity prompts have significant advantage across all three representative datasets. The diversity prompts mitigate the concentration of the generated reference image distribution, thereby enhancing its alignment with the Gaussian distribution of the test category.

\textbf{Stronger Diffusion, Stronger Classification} We further evaluate the zero-shot classification performance of various diffusion models. In terms of text-to-image generation quality, the models rank as SD-turbo > SD 1.5 > SD 1.3 > SD 1.1. Due to experimental time constraints, the number of reference images in this evaluation is set to 30. The results are shown in Figure \ref{fig:gdc_stronger_diffusion}. As depicted, diffusion models with superior generation performance also demonstrate significantly better classification performance. This finding underscores a key advantage of our method: stronger diffusion models lead to better classification results. Diffusion models can be improved through various approaches such as increasing data and adjusting the structure ~\citep{liu2024alignment, rombach2022high}, obtaining better initial noise~\citep{guo2024initno, qi2024not, zhou2024golden}, optimizing the sampling strategy~\citep{shao2024iv, lu2022dpm, lu2022dpm++, bai2024zigzagdiffusionsamplingpath}, and enhancing alignment capabilities~\citep{liu2024alignment, wallace2024diffusion, prabhudesai2023aligning}. Such advancements not only boost generative performance but also naturally enhance zero-shot classification capabilities.



\section{Discussion}
\label{sec:discuss}

\begin{figure}[th]
\centering
\includegraphics[width =1.0\linewidth ]{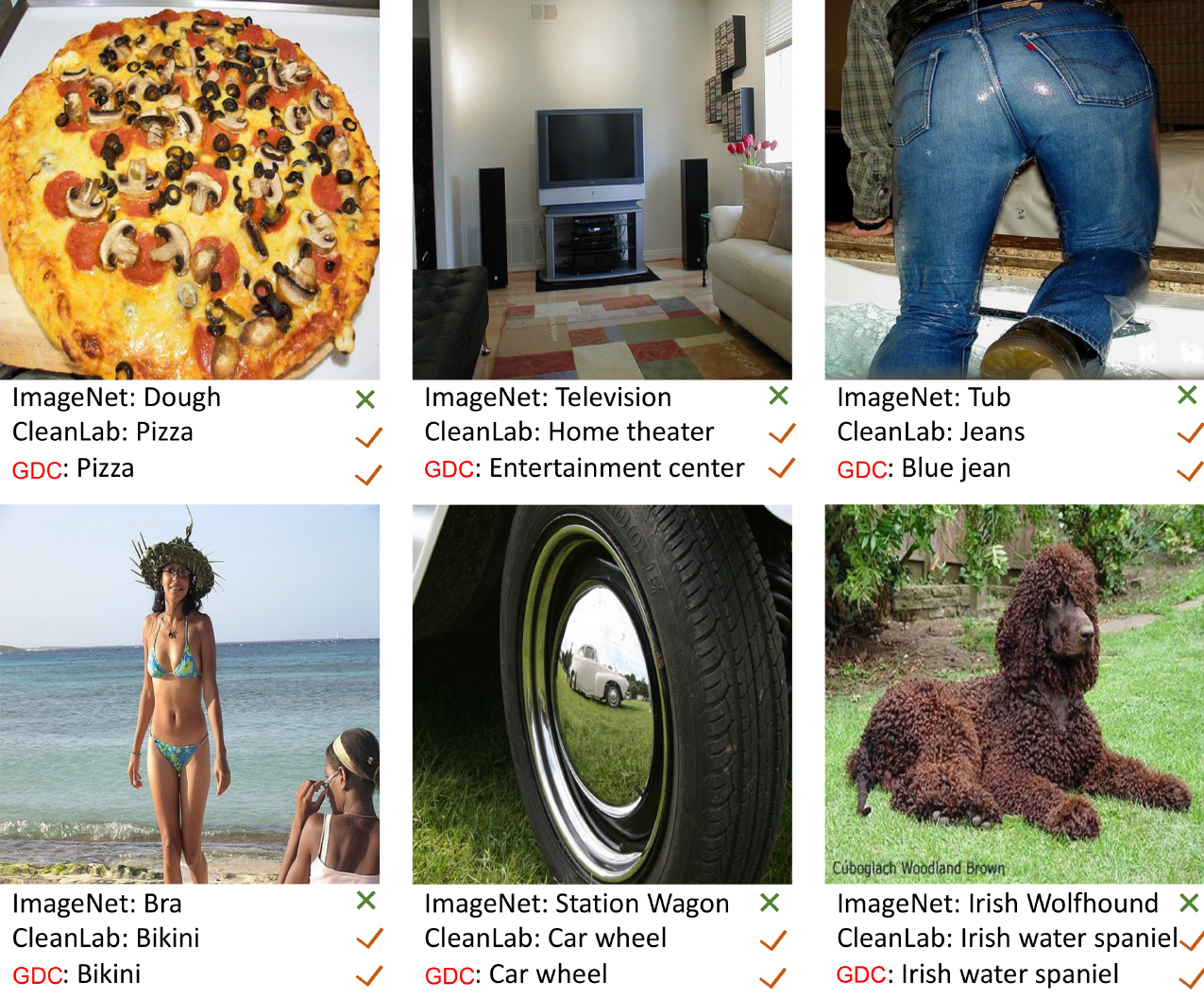}  
\caption{GDC can correct some common label errors in ImageNet. We provide examples illustrating instances where the ImageNet dataset assigns incorrect labels but GDC can annotate correctly. Cleanlab \citep{northcutt2021pervasive}, which focuses on correcting incorrect labels, can be accessed from this \href{https://github.com/cleanlab/cleanlab}{URL}. The results demonstrate that GDC yields equivalent correct results as Cleanlab for these hard cases and excels in distinguishing visually similar instances.}
\label{fig:wronglab} 
\end{figure}

In this section, we discuss case study, error analysis, limitations, and potential directions.

\textbf{Case Study} Through good and bad case analysis, we may highlight some success or failure modes of GDC. Recent studies analyzed label errors in ImageNet and pointed out that most label errors come from ambiguous instances and the resulting human uncertainty \citep{northcutt2021pervasive}. We particularly evaluate GDC on the instances difficult for human annotators and surprisingly observe that GDC is able to correctly classify many instances that human annotators cannot correctly classify. In Figure \ref{fig:wronglab}, we randomly present some instances in which ImageNet incorrectly annotates given by Cleanlab \citep{northcutt2021pervasive}. Fortunately, GDC yields reasonably good semantic labels for the shown cases.

\begin{figure}[h]
    \centering
    \includegraphics[width=1.1\linewidth]{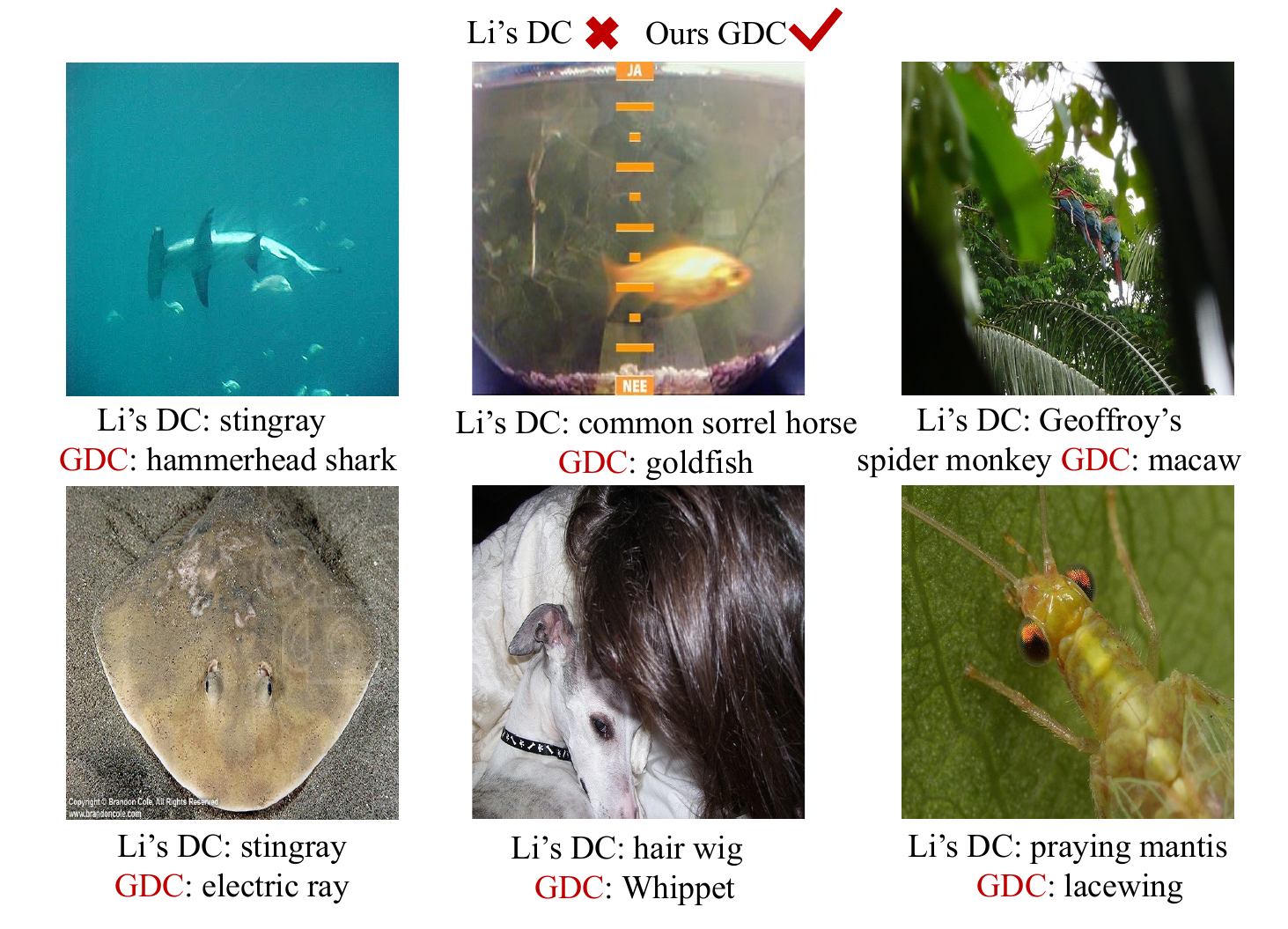}
    \caption{Some visual results misclassified by Li's DC are correctly classified by our GDC.}
    \label{fig:samples}
\end{figure}

\textbf{Visual Comparision}
In Figure \ref{fig:samples}, we illustrate visual examples where Li's DC misclassifies, but our GDC successfully classifies them. Our method shows exceptional performance in challenging scenarios, such as when the \textit{subject's texture is similar to the background} (e.g., hammerhead shark), \textit{the subject appears small} (e.g., macaw), or \textit{the subject is partially occluded} (e.g., goldfish and Whippet). These improvements can be credited to the diverse reference samples generated by SD, the critical features extracted by DINO, and the robust modeling power of GMM. These results inspire further research into Gaussian-based diffusion classification approaches.

\begin{figure}[h]
\centering
\includegraphics[width =1.0\linewidth ]{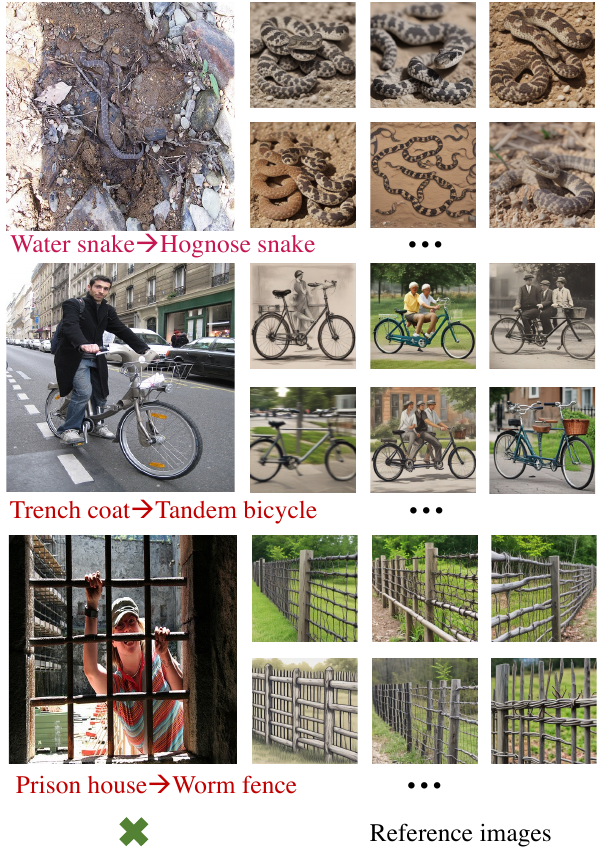}  
\caption{Correct Cases and Erroneous Cases. We show examples of correctly and incorrectly classified instances from ImageNet. The large images are the test images, while the small images are the generated reference images corresponding to the category assigned by GDC.}
\label{fig:case} 
\end{figure}

\textbf{Error Analysis} We first show multiple examples of correctly/incorrectly classified instances from ImageNet in Figure \ref{fig:case} and summarize three kinds of typical errors that may help us better understand and improve GDC. We try to understand when and why GDC fails. \textbf{First}, a class of erroneous cases fools GDC because the object is well hidden in the environment and is visually difficult for object recognition. For example, the misclassified water snake in the Figure \ref{fig:case} belong to this class of erroneous cases. In these cases, GDC may be misled by the environment but not the main semantic object. \textbf{Second}, another class of erroneous cases fool GDC because multiple semantic objects co-exist in the images. For example, the misclassified trench coat in the Figure \ref{fig:case} appears with a man riding a bicycle. In these cases, even human annotators may not exactly identify which is the main semantic label. We do not have a single ground-truth label for co-existent semantic objects. So, it actually makes sense that GDC predicts the semantic label of one of the co-existent semantic objects. \textbf{Third}, some erroneous cases fool GDC because the camera views are unusual. For example, the misclassified prison house in the Figure \ref{fig:case} is shown from the inside angle with a man standing outside of the prison cell. As for these cases, diffusion models rarely generate unusual camera views without specific prompts. To mitigate the typical errors, we believe improving the generation diversity and focusing on the main objects can be valuable. For example, we may design more strategies to improve the diversity of 1) background, 2) co-existent semantic objects, and 3) camera views. 

\textbf{Limitations} (1) While our method achieves impressive accuracy, GDC also regularly produces bad classification results for some typical cases. We do not completely know about the scope of these erroneous cases. This may limit the application scope of GDC for some settings. (2) We use image features with full dimensions to construct GMM. There may be redundancies and outliers here. Projecting the features into a low-dimensional space may a useful way to further analyze and enhance GDC. 

\textbf{Future Directions} (1) The framework of Gaussian-based generative classification be easily generalized to other learning tasks beyond standard classification, including novel class discovery and anomaly detection, particularly when obtaining training data is difficult. (2) It is also interesting to employ GDC-based methods to solve various domain tasks, such as medical image analysis and remote sensing, which may leverage the power of large diffusion models. (3) According to our observation that stronger diffusion models lead to stronger classifiers, zero-shot classification accuracy on ImageNet can naturally become a useful performance measure for text-to-image generation. It will be interesting to deeply analysis how generative abilities associate with the generated image embeddings.

\section{Conclusion}
\label{sec:conclusion}

In this paper, we aim at solving the efficiency bottleneck of zero-shot diffusion-based classifiers while maintaining good accuracy. Leveraging the powerful generative abilities of pretrained diffusion models without requiring any training data, the proposed GDC not only well solved the extremely poor efficiency problem but also achieved very competitive classification performance on popular benchmark datasets. For example, on ImageNet, the most popular classification dataset, GDC accelerates single-image classification by 30000 times and outperforms previous diffusion-based classifiers by 10 points. As the first zero-shot diffusion-based classifier that exhibits both competitive accuracy and practical efficiency, our work makes zero-shot diffusion-based classifiers get much closer to real-world applications. With stronger diffusion models that will appear in future, GDC can even naturally self-improve these advantages. While some limitations still exist, it will be very promising to further understanding and improve this approach.


{
    \small
    \bibliographystyle{ieeenat_fullname}
    \bibliography{main}

\begin{thebibliography}{75}
\providecommand{\natexlab}[1]{#1}
\providecommand{\url}[1]{\texttt{#1}}
\expandafter\ifx\csname urlstyle\endcsname\relax
  \providecommand{\doi}[1]{doi: #1}\else
  \providecommand{\doi}{doi: \begingroup \urlstyle{rm}\Url}\fi

\bibitem[Bai et~al.(2024)Bai, Shao, Zhou, Qi, Xu, Xiong, and Xie]{bai2024zigzagdiffusionsamplingpath}
Lichen Bai, Shitong Shao, Zikai Zhou, Zipeng Qi, Zhiqiang Xu, Haoyi Xiong, and Zeke Xie.
\newblock Zigzag diffusion sampling: The path to success is zigzag, 2024.

\bibitem[Benoit(1924)]{benoit1924note}
Commandant Benoit.
\newblock Note sur une m{\'e}thode de r{\'e}solution des {\'e}quations normales provenant de l'application de la m{\'e}thode des moindres carr{\'e}s {\`a} un syst{\`e}me d'{\'e}quations lin{\'e}aires en nombre inf{\'e}rieur {\`a} celui des inconnues (proc{\'e}d{\'e} du commandant cholesky).
\newblock \emph{Bulletin g{\'e}od{\'e}sique}, 2\penalty0 (1):\penalty0 67--77, 1924.

\bibitem[Blattmann et~al.(2023)Blattmann, Dockhorn, Kulal, Mendelevitch, Kilian, Lorenz, Levi, English, Voleti, Letts, et~al.]{blattmann2023stable}
Andreas Blattmann, Tim Dockhorn, Sumith Kulal, Daniel Mendelevitch, Maciej Kilian, Dominik Lorenz, Yam Levi, Zion English, Vikram Voleti, Adam Letts, et~al.
\newblock Stable video diffusion: Scaling latent video diffusion models to large datasets.
\newblock \emph{arXiv preprint arXiv:2311.15127}, 2023.

\bibitem[Bossard et~al.(2014)Bossard, Guillaumin, and Van~Gool]{bossard2014food}
Lukas Bossard, Matthieu Guillaumin, and Luc Van~Gool.
\newblock Food-101--mining discriminative components with random forests.
\newblock In \emph{Computer Vision--ECCV 2014: 13th European Conference, Zurich, Switzerland, September 6-12, 2014, Proceedings, Part VI 13}, pages 446--461. Springer, 2014.

\bibitem[Brown et~al.(2020)Brown, Mann, Ryder, Subbiah, Kaplan, Dhariwal, Neelakantan, Shyam, Sastry, Askell, et~al.]{brown2020language}
Tom Brown, Benjamin Mann, Nick Ryder, Melanie Subbiah, Jared~D Kaplan, Prafulla Dhariwal, Arvind Neelakantan, Pranav Shyam, Girish Sastry, Amanda Askell, et~al.
\newblock Language models are few-shot learners.
\newblock \emph{Advances in neural information processing systems}, 33:\penalty0 1877--1901, 2020.

\bibitem[Bucher et~al.(2017)Bucher, Herbin, and Jurie]{bucher2017generating}
Maxime Bucher, St{\'e}phane Herbin, and Fr{\'e}d{\'e}ric Jurie.
\newblock Generating visual representations for zero-shot classification.
\newblock In \emph{Proceedings of the IEEE International Conference on Computer Vision Workshops}, pages 2666--2673, 2017.

\bibitem[Burgert et~al.(2022)Burgert, Ranasinghe, Li, and Ryoo]{burgert2022peekaboo}
Ryan Burgert, Kanchana Ranasinghe, Xiang Li, and Michael~S Ryoo.
\newblock Peekaboo: Text to image diffusion models are zero-shot segmentors.
\newblock \emph{arXiv preprint arXiv:2211.13224}, 2022.

\bibitem[Chen et~al.(2023)Chen, Dong, Wang, Yang, Duan, Su, and Zhu]{chen2023robust}
Huanran Chen, Yinpeng Dong, Zhengyi Wang, Xiao Yang, Chengqi Duan, Hang Su, and Jun Zhu.
\newblock Robust classification via a single diffusion model.
\newblock \emph{arXiv preprint arXiv:2305.15241}, 2023.

\bibitem[Cimpoi et~al.(2014)Cimpoi, Maji, Kokkinos, Mohamed, , and Vedaldi]{cimpoi14describing}
M. Cimpoi, S. Maji, I. Kokkinos, S. Mohamed, , and A. Vedaldi.
\newblock Describing textures in the wild.
\newblock In \emph{Proceedings of the {IEEE} Conf. on Computer Vision and Pattern Recognition ({CVPR})}, 2014.

\bibitem[Clark and Jaini(2023)]{clark2023text}
Kevin Clark and Priyank Jaini.
\newblock Text-to-image diffusion models are zero-shot classifiers.
\newblock In \emph{ICLR 2023 Workshop on Mathematical and Empirical Understanding of Foundation Models}, 2023.

\bibitem[Coates et~al.(2011)Coates, Ng, and Lee]{coates2011analysis}
Adam Coates, Andrew Ng, and Honglak Lee.
\newblock An analysis of single-layer networks in unsupervised feature learning.
\newblock In \emph{Proceedings of the fourteenth international conference on artificial intelligence and statistics}, pages 215--223. JMLR Workshop and Conference Proceedings, 2011.

\bibitem[Croce et~al.(2020)Croce, Castellucci, and Basili]{croce2020gan}
Danilo Croce, Giuseppe Castellucci, and Roberto Basili.
\newblock Gan-bert: Generative adversarial learning for robust text classification with a bunch of labeled examples.
\newblock In \emph{Proceedings of the 58th Annual Meeting of the Association for Computational Linguistics}, pages 2114--2119, 2020.

\bibitem[Deng et~al.(2009)Deng, Dong, Socher, Li, Li, and Fei-Fei]{deng2009ImageNet}
Jia Deng, Wei Dong, Richard Socher, Li-Jia Li, Kai Li, and Li Fei-Fei.
\newblock Imagenet: A large-scale hierarchical image database.
\newblock In \emph{2009 IEEE conference on computer vision and pattern recognition}, pages 248--255. Ieee, 2009.

\bibitem[Dhariwal and Nichol(2021)]{dhariwal2021diffusion}
Prafulla Dhariwal and Alexander Nichol.
\newblock Diffusion models beat gans on image synthesis.
\newblock \emph{Advances in neural information processing systems}, 34:\penalty0 8780--8794, 2021.

\bibitem[Du~Plessis and Sugiyama(2014)]{du2014class}
Marthinus~Christoffel Du~Plessis and Masashi Sugiyama.
\newblock Class prior estimation from positive and unlabeled data.
\newblock \emph{IEICE TRANSACTIONS on Information and Systems}, 97\penalty0 (5):\penalty0 1358--1362, 2014.

\bibitem[Fei-Fei et~al.(2006)Fei-Fei, Fergus, and Perona]{fei2006one}
Li Fei-Fei, Robert Fergus, and Pietro Perona.
\newblock One-shot learning of object categories.
\newblock \emph{IEEE transactions on pattern analysis and machine intelligence}, 28\penalty0 (4):\penalty0 594--611, 2006.

\bibitem[Guo et~al.(2024)Guo, Liu, Cui, Li, Yang, and Huang]{guo2024initno}
Xiefan Guo, Jinlin Liu, Miaomiao Cui, Jiankai Li, Hongyu Yang, and Di Huang.
\newblock Initno: Boosting text-to-image diffusion models via initial noise optimization.
\newblock In \emph{Proceedings of the IEEE/CVF Conference on Computer Vision and Pattern Recognition}, pages 9380--9389, 2024.

\bibitem[He et~al.(2022)He, Chen, Xie, Li, Doll{\'a}r, and Girshick]{he2022masked}
Kaiming He, Xinlei Chen, Saining Xie, Yanghao Li, Piotr Doll{\'a}r, and Ross Girshick.
\newblock Masked autoencoders are scalable vision learners.
\newblock In \emph{Proceedings of the IEEE/CVF conference on computer vision and pattern recognition}, pages 16000--16009, 2022.

\bibitem[Hjelm et~al.(2018)Hjelm, Fedorov, Lavoie-Marchildon, Grewal, Bachman, Trischler, and Bengio]{hjelm2018learning}
R~Devon Hjelm, Alex Fedorov, Samuel Lavoie-Marchildon, Karan Grewal, Phil Bachman, Adam Trischler, and Yoshua Bengio.
\newblock Learning deep representations by mutual information estimation and maximization.
\newblock In \emph{International Conference on Learning Representations}, 2018.

\bibitem[Ho et~al.(2020)Ho, Jain, and Abbeel]{ho2020denoising}
Jonathan Ho, Ajay Jain, and Pieter Abbeel.
\newblock Denoising diffusion probabilistic models.
\newblock \emph{Advances in neural information processing systems}, 33:\penalty0 6840--6851, 2020.

\bibitem[Ho et~al.(2022)Ho, Chan, Saharia, Whang, Gao, Gritsenko, Kingma, Poole, Norouzi, Fleet, et~al.]{ho2022imagen}
Jonathan Ho, William Chan, Chitwan Saharia, Jay Whang, Ruiqi Gao, Alexey Gritsenko, Diederik~P Kingma, Ben Poole, Mohammad Norouzi, David~J Fleet, et~al.
\newblock Imagen video: High definition video generation with diffusion models.
\newblock \emph{arXiv preprint arXiv:2210.02303}, 2022.

\bibitem[Jaini et~al.(2023)Jaini, Clark, and Geirhos]{jaini2023intriguing}
Priyank Jaini, Kevin Clark, and Robert Geirhos.
\newblock Intriguing properties of generative classifiers.
\newblock In \emph{The Twelfth International Conference on Learning Representations}, 2023.

\bibitem[Jiang et~al.(2017)Jiang, Wang, Shan, Yang, and Chen]{jiang2017learning}
Huajie Jiang, Ruiping Wang, Shiguang Shan, Yi Yang, and Xilin Chen.
\newblock Learning discriminative latent attributes for zero-shot classification.
\newblock In \emph{Proceedings of the IEEE International Conference on Computer Vision}, pages 4223--4232, 2017.

\bibitem[Kong et~al.(2020)Kong, Ping, Huang, Zhao, and Catanzaro]{kong2020diffwave}
Zhifeng Kong, Wei Ping, Jiaji Huang, Kexin Zhao, and Bryan Catanzaro.
\newblock Diffwave: A versatile diffusion model for audio synthesis.
\newblock In \emph{International Conference on Learning Representations}, 2020.

\bibitem[Krizhevsky and Hinton(2009)]{krizhevsky2009learning}
Alex Krizhevsky and Geoffrey Hinton.
\newblock Learning multiple layers of features from tiny images.
\newblock 2009.

\bibitem[Lawrence et~al.(2002)Lawrence, Burns, Back, Tsoi, and Giles]{lawrence2002neural}
Steve Lawrence, Ian Burns, Andrew Back, Ah~Chung Tsoi, and C~Lee Giles.
\newblock Neural network classification and prior class probabilities.
\newblock In \emph{Neural networks: tricks of the trade}, pages 299--313. Springer, 2002.

\bibitem[Lee et~al.(2019)Lee, Yun, Lee, Lee, Li, and Shin]{lee2019robust}
Kimin Lee, Sukmin Yun, Kibok Lee, Honglak Lee, Bo Li, and Jinwoo Shin.
\newblock Robust inference via generative classifiers for handling noisy labels.
\newblock In \emph{International conference on machine learning}, pages 3763--3772. PMLR, 2019.

\bibitem[Li et~al.(2023)Li, Prabhudesai, Duggal, Brown, and Pathak]{li2023diffusion}
Alexander~C. Li, Mihir Prabhudesai, Shivam Duggal, Ellis Brown, and Deepak Pathak.
\newblock Your diffusion model is secretly a zero-shot classifier.
\newblock In \emph{Thirty-seventh International Conference on Computer Vision}, 2023.

\bibitem[Li et~al.(2019)Li, Bradshaw, and Sharma]{li2019generative}
Yingzhen Li, John Bradshaw, and Yash Sharma.
\newblock Are generative classifiers more robust to adversarial attacks?
\newblock In \emph{International Conference on Machine Learning}, pages 3804--3814. PMLR, 2019.

\bibitem[Lin et~al.(2023)Lin, Gao, Tang, Takikawa, Zeng, Huang, Kreis, Fidler, Liu, and Lin]{lin2023magic3d}
Chen-Hsuan Lin, Jun Gao, Luming Tang, Towaki Takikawa, Xiaohui Zeng, Xun Huang, Karsten Kreis, Sanja Fidler, Ming-Yu Liu, and Tsung-Yi Lin.
\newblock Magic3d: High-resolution text-to-3d content creation.
\newblock In \emph{Proceedings of the IEEE/CVF Conference on Computer Vision and Pattern Recognition}, pages 300--309, 2023.

\bibitem[Liu et~al.(2024)Liu, Shao, Li, Bai, Xu, Xiong, Kwok, Helal, and Xie]{liu2024alignment}
Buhua Liu, Shitong Shao, Bao Li, Lichen Bai, Zhiqiang Xu, Haoyi Xiong, James Kwok, Sumi Helal, and Zeke Xie.
\newblock Alignment of diffusion models: Fundamentals, challenges, and future.
\newblock \emph{arXiv preprint arXiv:2409.07253}, 2024.

\bibitem[Lu et~al.(2022{\natexlab{a}})Lu, Zhou, Bao, Chen, Li, and Zhu]{lu2022dpm}
Cheng Lu, Yuhao Zhou, Fan Bao, Jianfei Chen, Chongxuan Li, and Jun Zhu.
\newblock Dpm-solver: A fast ode solver for diffusion probabilistic model sampling in around 10 steps.
\newblock \emph{Advances in Neural Information Processing Systems}, 35:\penalty0 5775--5787, 2022{\natexlab{a}}.

\bibitem[Lu et~al.(2022{\natexlab{b}})Lu, Zhou, Bao, Chen, Li, and Zhu]{lu2022dpm++}
Cheng Lu, Yuhao Zhou, Fan Bao, Jianfei Chen, Chongxuan Li, and Jun Zhu.
\newblock Dpm-solver++: Fast solver for guided sampling of diffusion probabilistic models.
\newblock \emph{arXiv preprint arXiv:2211.01095}, 2022{\natexlab{b}}.

\bibitem[Mackowiak et~al.(2021)Mackowiak, Ardizzone, Kothe, and Rother]{mackowiak2021generative}
Radek Mackowiak, Lynton Ardizzone, Ullrich Kothe, and Carsten Rother.
\newblock Generative classifiers as a basis for trustworthy image classification.
\newblock In \emph{Proceedings of the IEEE/CVF Conference on Computer Vision and Pattern Recognition}, pages 2971--2981, 2021.

\bibitem[Mensink et~al.(2014)Mensink, Gavves, and Snoek]{mensink2014costa}
Thomas Mensink, Efstratios Gavves, and Cees~GM Snoek.
\newblock Costa: Co-occurrence statistics for zero-shot classification.
\newblock In \emph{Proceedings of the IEEE conference on computer vision and pattern recognition}, pages 2441--2448, 2014.

\bibitem[Ng and Jordan(2001)]{ng2001discriminative}
Andrew Ng and Michael Jordan.
\newblock On discriminative vs. generative classifiers: A comparison of logistic regression and naive bayes.
\newblock \emph{Advances in neural information processing systems}, 14, 2001.

\bibitem[Nichol et~al.(2021)Nichol, Dhariwal, Ramesh, Shyam, Mishkin, McGrew, Sutskever, and Chen]{nichol2021glide}
Alex Nichol, Prafulla Dhariwal, Aditya Ramesh, Pranav Shyam, Pamela Mishkin, Bob McGrew, Ilya Sutskever, and Mark Chen.
\newblock Glide: Towards photorealistic image generation and editing with text-guided diffusion models.
\newblock \emph{arXiv preprint arXiv:2112.10741}, 2021.

\bibitem[Nichol et~al.(2022)Nichol, Dhariwal, Ramesh, Shyam, Mishkin, Mcgrew, Sutskever, and Chen]{nichol2022glide}
Alexander~Quinn Nichol, Prafulla Dhariwal, Aditya Ramesh, Pranav Shyam, Pamela Mishkin, Bob Mcgrew, Ilya Sutskever, and Mark Chen.
\newblock Glide: Towards photorealistic image generation and editing with text-guided diffusion models.
\newblock In \emph{International Conference on Machine Learning}, pages 16784--16804. PMLR, 2022.

\bibitem[Nilsback and Zisserman(2008)]{nilsback2008automated}
Maria-Elena Nilsback and Andrew Zisserman.
\newblock Automated flower classification over a large number of classes.
\newblock In \emph{2008 Sixth Indian conference on computer vision, graphics \& image processing}, pages 722--729. IEEE, 2008.

\bibitem[Northcutt et~al.(2021)Northcutt, Athalye, and Mueller]{northcutt2021pervasive}
Curtis~G Northcutt, Anish Athalye, and Jonas Mueller.
\newblock Pervasive label errors in test sets destabilize machine learning benchmarks.
\newblock In \emph{Thirty-fifth Conference on Neural Information Processing Systems Datasets and Benchmarks Track (Round 1)}, 2021.

\bibitem[Oquab et~al.(2023)Oquab, Darcet, Moutakanni, Vo, Szafraniec, Khalidov, Fernandez, Haziza, Massa, El-Nouby, et~al.]{oquab2023dinov2}
Maxime Oquab, Timoth{\'e}e Darcet, Th{\'e}o Moutakanni, Huy Vo, Marc Szafraniec, Vasil Khalidov, Pierre Fernandez, Daniel Haziza, Francisco Massa, Alaaeldin El-Nouby, et~al.
\newblock Dinov2: Learning robust visual features without supervision.
\newblock \emph{arXiv preprint arXiv:2304.07193}, 2023.

\bibitem[Osowiechi et~al.(2024)Osowiechi, Noori, Hakim, Yazdanpanah, Bahri, Cheraghalikhani, Dastani, Beizaee, Ayed, and Desrosiers]{osowiechi2024watt}
David Osowiechi, Mehrdad Noori, Gustavo Adolfo~Vargas Hakim, Moslem Yazdanpanah, Ali Bahri, Milad Cheraghalikhani, Sahar Dastani, Farzad Beizaee, Ismail~Ben Ayed, and Christian Desrosiers.
\newblock Watt: Weight average test-time adaption of clip.
\newblock \emph{arXiv preprint arXiv:2406.13875}, 2024.

\bibitem[Parkhi et~al.(2012)Parkhi, Vedaldi, Zisserman, and Jawahar]{parkhi2012cats}
Omkar~M Parkhi, Andrea Vedaldi, Andrew Zisserman, and CV Jawahar.
\newblock Cats and dogs.
\newblock In \emph{2012 IEEE conference on computer vision and pattern recognition}, pages 3498--3505. IEEE, 2012.

\bibitem[Peebles and Xie(2023)]{peebles2023scalable}
William Peebles and Saining Xie.
\newblock Scalable diffusion models with transformers.
\newblock In \emph{Proceedings of the IEEE/CVF International Conference on Computer Vision}, pages 4195--4205, 2023.

\bibitem[Podell et~al.(2023)Podell, English, Lacey, Blattmann, Dockhorn, M{\"u}ller, Penna, and Rombach]{podell2023sdxl}
Dustin Podell, Zion English, Kyle Lacey, Andreas Blattmann, Tim Dockhorn, Jonas M{\"u}ller, Joe Penna, and Robin Rombach.
\newblock Sdxl: Improving latent diffusion models for high-resolution image synthesis.
\newblock \emph{arXiv preprint arXiv:2307.01952}, 2023.

\bibitem[Poole et~al.(2022)Poole, Jain, Barron, and Mildenhall]{poole2022dreamfusion}
Ben Poole, Ajay Jain, Jonathan~T Barron, and Ben Mildenhall.
\newblock Dreamfusion: Text-to-3d using 2d diffusion.
\newblock In \emph{The Eleventh International Conference on Learning Representations}, 2022.

\bibitem[Prabhudesai et~al.(2023)Prabhudesai, Goyal, Pathak, and Fragkiadaki]{prabhudesai2023aligning}
Mihir Prabhudesai, Anirudh Goyal, Deepak Pathak, and Katerina Fragkiadaki.
\newblock Aligning text-to-image diffusion models with reward backpropagation.
\newblock \emph{arXiv preprint arXiv:2310.03739}, 2023.

\bibitem[Qi et~al.(2023{\natexlab{a}})Qi, Huang, Huang, Guo, Chen, Han, Wang, Zhang, Liu, Ding, et~al.]{qi2023layered}
Zipeng Qi, Guoxi Huang, Zebin Huang, Qin Guo, Jinwen Chen, Junyu Han, Jian Wang, Gang Zhang, Lufei Liu, Errui Ding, et~al.
\newblock Layered rendering diffusion model for zero-shot guided image synthesis.
\newblock \emph{arXiv preprint arXiv:2311.18435}, 2023{\natexlab{a}}.

\bibitem[Qi et~al.(2023{\natexlab{b}})Qi, Zhang, Cheng, Xiao, and Wang]{qi2023difftalker}
Zipeng Qi, Xulong Zhang, Ning Cheng, Jing Xiao, and Jianzong Wang.
\newblock Difftalker: Co-driven audio-image diffusion for talking faces via intermediate landmarks.
\newblock \emph{arXiv preprint arXiv:2309.07509}, 2023{\natexlab{b}}.

\bibitem[Qi et~al.(2024)Qi, Bai, Xiong, et~al.]{qi2024not}
Zipeng Qi, Lichen Bai, Haoyi Xiong, et~al.
\newblock Not all noises are created equally: Diffusion noise selection and optimization.
\newblock \emph{arXiv preprint arXiv:2407.14041}, 2024.

\bibitem[Radford et~al.(2019)Radford, Wu, Child, Luan, Amodei, Sutskever, et~al.]{radford2019language}
Alec Radford, Jeffrey Wu, Rewon Child, David Luan, Dario Amodei, Ilya Sutskever, et~al.
\newblock Language models are unsupervised multitask learners.
\newblock \emph{OpenAI blog}, 1\penalty0 (8):\penalty0 9, 2019.

\bibitem[Radford et~al.(2021)Radford, Kim, Hallacy, Ramesh, Goh, Agarwal, Sastry, Askell, Mishkin, Clark, et~al.]{radford2021learning}
Alec Radford, Jong~Wook Kim, Chris Hallacy, Aditya Ramesh, Gabriel Goh, Sandhini Agarwal, Girish Sastry, Amanda Askell, Pamela Mishkin, Jack Clark, et~al.
\newblock Learning transferable visual models from natural language supervision.
\newblock In \emph{International conference on machine learning}, pages 8748--8763. PMLR, 2021.

\bibitem[Ramesh et~al.(2021)Ramesh, Pavlov, Goh, Gray, Voss, Radford, Chen, and Sutskever]{ramesh2021zero}
Aditya Ramesh, Mikhail Pavlov, Gabriel Goh, Scott Gray, Chelsea Voss, Alec Radford, Mark Chen, and Ilya Sutskever.
\newblock Zero-shot text-to-image generation.
\newblock In \emph{International Conference on Machine Learning}, pages 8821--8831. PMLR, 2021.

\bibitem[Ramesh et~al.(2022)Ramesh, Dhariwal, Nichol, Chu, and Chen]{ramesh2022hierarchical}
Aditya Ramesh, Prafulla Dhariwal, Alex Nichol, Casey Chu, and Mark Chen.
\newblock Hierarchical text-conditional image generation with clip latents.
\newblock \emph{arXiv preprint arXiv:2204.06125}, 1\penalty0 (2):\penalty0 3, 2022.

\bibitem[Rombach et~al.(2022)Rombach, Blattmann, Lorenz, Esser, and Ommer]{rombach2022high}
Robin Rombach, Andreas Blattmann, Dominik Lorenz, Patrick Esser, and Bj{\"o}rn Ommer.
\newblock High-resolution image synthesis with latent diffusion models.
\newblock In \emph{Proceedings of the IEEE/CVF conference on computer vision and pattern recognition}, pages 10684--10695, 2022.

\bibitem[Saharia et~al.(2022)Saharia, Chan, Saxena, Li, Whang, Denton, Ghasemipour, Gontijo~Lopes, Karagol~Ayan, Salimans, et~al.]{saharia2022photorealistic}
Chitwan Saharia, William Chan, Saurabh Saxena, Lala Li, Jay Whang, Emily~L Denton, Kamyar Ghasemipour, Raphael Gontijo~Lopes, Burcu Karagol~Ayan, Tim Salimans, et~al.
\newblock Photorealistic text-to-image diffusion models with deep language understanding.
\newblock \emph{Advances in neural information processing systems}, 35:\penalty0 36479--36494, 2022.

\bibitem[Sauer et~al.(2023)Sauer, Lorenz, Blattmann, and Rombach]{sauer2023adversarial}
Axel Sauer, Dominik Lorenz, Andreas Blattmann, and Robin Rombach.
\newblock Adversarial diffusion distillation.
\newblock \emph{arXiv preprint arXiv:2311.17042}, 2023.

\bibitem[Sensoy et~al.(2020)Sensoy, Kaplan, Cerutti, and Saleki]{sensoy2020uncertainty}
Murat Sensoy, Lance Kaplan, Federico Cerutti, and Maryam Saleki.
\newblock Uncertainty-aware deep classifiers using generative models.
\newblock In \emph{Proceedings of the AAAI conference on artificial intelligence}, pages 5620--5627, 2020.

\bibitem[Shao et~al.(2024)Shao, Zhou, Bai, Xiong, and Xie]{shao2024iv}
Shitong Shao, Zikai Zhou, Lichen Bai, Haoyi Xiong, and Zeke Xie.
\newblock Iv-mixed sampler: Leveraging image diffusion models for enhanced video synthesis.
\newblock \emph{arXiv preprint arXiv:2410.04171}, 2024.

\bibitem[Shapiro and Wilk(1965)]{shapiro1965analysis}
Samuel~Sanford Shapiro and Martin~B Wilk.
\newblock An analysis of variance test for normality (complete samples).
\newblock \emph{Biometrika}, 52\penalty0 (3-4):\penalty0 591--611, 1965.

\bibitem[Sohl-Dickstein et~al.(2015)Sohl-Dickstein, Weiss, Maheswaranathan, and Ganguli]{sohl2015deep}
Jascha Sohl-Dickstein, Eric Weiss, Niru Maheswaranathan, and Surya Ganguli.
\newblock Deep unsupervised learning using nonequilibrium thermodynamics.
\newblock In \emph{International conference on machine learning}, pages 2256--2265. PMLR, 2015.

\bibitem[Song and Ermon(2019)]{song2019generative}
Yang Song and Stefano Ermon.
\newblock Generative modeling by estimating gradients of the data distribution.
\newblock \emph{Advances in neural information processing systems}, 32, 2019.

\bibitem[Van De~Ven et~al.(2021)Van De~Ven, Li, and Tolias]{van2021class}
Gido~M Van De~Ven, Zhe Li, and Andreas~S Tolias.
\newblock Class-incremental learning with generative classifiers.
\newblock In \emph{Proceedings of the IEEE/CVF Conference on Computer Vision and Pattern Recognition}, pages 3611--3620, 2021.

\bibitem[Wallace et~al.(2024)Wallace, Dang, Rafailov, Zhou, Lou, Purushwalkam, Ermon, Xiong, Joty, and Naik]{wallace2024diffusion}
Bram Wallace, Meihua Dang, Rafael Rafailov, Linqi Zhou, Aaron Lou, Senthil Purushwalkam, Stefano Ermon, Caiming Xiong, Shafiq Joty, and Nikhil Naik.
\newblock Diffusion model alignment using direct preference optimization.
\newblock In \emph{Proceedings of the IEEE/CVF Conference on Computer Vision and Pattern Recognition}, pages 8228--8238, 2024.

\bibitem[Wang et~al.(2019)Wang, Zheng, Yu, and Miao]{wang2019survey}
Wei Wang, Vincent~W Zheng, Han Yu, and Chunyan Miao.
\newblock A survey of zero-shot learning: Settings, methods, and applications.
\newblock \emph{ACM Transactions on Intelligent Systems and Technology (TIST)}, 10\penalty0 (2):\penalty0 1--37, 2019.

\bibitem[Wei et~al.(2021)Wei, Bosma, Zhao, Guu, Yu, Lester, Du, Dai, and Le]{wei2021finetuned}
Jason Wei, Maarten Bosma, Vincent~Y Zhao, Kelvin Guu, Adams~Wei Yu, Brian Lester, Nan Du, Andrew~M Dai, and Quoc~V Le.
\newblock Finetuned language models are zero-shot learners.
\newblock \emph{arXiv preprint arXiv:2109.01652}, 2021.

\bibitem[Wortsman et~al.(2022)Wortsman, Ilharco, Kim, Li, Kornblith, Roelofs, Lopes, Hajishirzi, Farhadi, Namkoong, et~al.]{wortsman2022robust}
Mitchell Wortsman, Gabriel Ilharco, Jong~Wook Kim, Mike Li, Simon Kornblith, Rebecca Roelofs, Raphael~Gontijo Lopes, Hannaneh Hajishirzi, Ali Farhadi, Hongseok Namkoong, et~al.
\newblock Robust fine-tuning of zero-shot models.
\newblock In \emph{Proceedings of the IEEE/CVF Conference on Computer Vision and Pattern Recognition}, pages 7959--7971, 2022.

\bibitem[Xian et~al.(2016)Xian, Akata, Sharma, Nguyen, Hein, and Schiele]{xian2016latent}
Yongqin Xian, Zeynep Akata, Gaurav Sharma, Quynh Nguyen, Matthias Hein, and Bernt Schiele.
\newblock Latent embeddings for zero-shot classification.
\newblock In \emph{Proceedings of the IEEE conference on computer vision and pattern recognition}, pages 69--77, 2016.

\bibitem[Xu et~al.(2023)Xu, Wang, Cheng, Cao, Shan, Qie, and Gao]{xu2023dream3d}
Jiale Xu, Xintao Wang, Weihao Cheng, Yan-Pei Cao, Ying Shan, Xiaohu Qie, and Shenghua Gao.
\newblock Dream3d: Zero-shot text-to-3d synthesis using 3d shape prior and text-to-image diffusion models.
\newblock In \emph{Proceedings of the IEEE/CVF Conference on Computer Vision and Pattern Recognition}, pages 20908--20918, 2023.

\bibitem[Yang et~al.(2023{\natexlab{a}})Yang, Yu, Wang, Wang, Weng, Zou, and Yu]{yang2023diffsound}
Dongchao Yang, Jianwei Yu, Helin Wang, Wen Wang, Chao Weng, Yuexian Zou, and Dong Yu.
\newblock Diffsound: Discrete diffusion model for text-to-sound generation.
\newblock \emph{IEEE/ACM Transactions on Audio, Speech, and Language Processing}, 2023{\natexlab{a}}.

\bibitem[Yang et~al.(2023{\natexlab{b}})Yang, Zhang, Song, Hong, Xu, Zhao, Zhang, Cui, and Yang]{yang2023diffusion}
Ling Yang, Zhilong Zhang, Yang Song, Shenda Hong, Runsheng Xu, Yue Zhao, Wentao Zhang, Bin Cui, and Ming-Hsuan Yang.
\newblock Diffusion models: A comprehensive survey of methods and applications.
\newblock \emph{ACM Computing Surveys}, 56\penalty0 (4):\penalty0 1--39, 2023{\natexlab{b}}.

\bibitem[Ye and Guo(2017)]{ye2017zero}
Meng Ye and Yuhong Guo.
\newblock Zero-shot classification with discriminative semantic representation learning.
\newblock In \emph{Proceedings of the IEEE conference on computer vision and pattern recognition}, pages 7140--7148, 2017.

\bibitem[Zheng et~al.(2023)Zheng, Wu, Bao, Cao, Li, and Zhu]{zheng2023revisiting}
Chenyu Zheng, Guoqiang Wu, Fan Bao, Yue Cao, Chongxuan Li, and Jun Zhu.
\newblock Revisiting discriminative vs. generative classifiers: Theory and implications.
\newblock In \emph{International Conference on Machine Learning}, 2023.

\bibitem[Zhou et~al.(2024)Zhou, Shao, Bai, Xu, Han, and Xie]{zhou2024golden}
Zikai Zhou, Shitong Shao, Lichen Bai, Zhiqiang Xu, Bo Han, and Zeke Xie.
\newblock Golden noise for diffusion models: A learning framework.
\newblock \emph{arXiv preprint arXiv:2411.09502}, 2024.

\bibitem[Zimmermann et~al.(2021)Zimmermann, Schott, Song, Dunn, and Klindt]{zimmermann2021score}
Roland~S Zimmermann, Lukas Schott, Yang Song, Benjamin~A Dunn, and David~A Klindt.
\newblock Score-based generative classifiers.
\newblock \emph{arXiv preprint arXiv:2110.00473}, 2021.

\end{thebibliography}
}

\clearpage
\appendix
\maketitlesupplementary

\section{Experimental Settings of Main Experiments}
\label{sec:expsetting}

\textbf{Computational environment.} The experiments are conducted on a computing cluster with GPUs of NVIDIA\textsuperscript{\textregistered} Tesla\textsuperscript{\texttrademark} V100, and Intel(R) Xeon(R) Platinum 8352V CPU @ 2.10GHz.

\subsection{Datasets and Data Preprocessing}
\label{sec:dataset}
For a comprehensive comparison, we conduct experiments across seven benchmark classification datasets as follows:

\textbf{ImageNet-1K}~\citep{deng2009ImageNet} is a large-scale visual database designed for visual object recognition research. It encompasses 1,000 distinct categories, with each category featuring 50 images in the validation subdataset. The challenges posed by this dataset for image classification include some categories with similar appearances and variations in resolutions across the dataset.

\textbf{CIFAR-10}/\textbf{CIFAR-100}~\citep{krizhevsky2009learning} includes 10/100 different categories, respectively, covering a range of common objects. Each dataset comprises 10,000 test images with dimensions of $32 \times 32$ pixels. The low resolution poses a challenge for the classification task.

\textbf{Flower-102}~\citep{nilsback2008automated} comprises 102 distinct flower categories, with each category containing between 40 and 258 images, all with resolutions exceeding $500 \times 500$ pixels. The test subdataset includes over 6,000 images. The images exhibit variations in scale, poses, and lighting. Moreover, certain categories display significant intra-category variations, and some categories are remarkably similar. These factors contribute to the increased difficulty of classification tasks.

\textbf{Oxford Pet-37}~\citep{parkhi2012cats} is a widely used pet classification dataset featuring 37 categories, each containing approximately 200 images. The test subdataset comprises over 3,600 images with varying resolutions.

\textbf{Food-101}~\citep{bossard2014food} is a challenging food recognition dataset with 101 categories. Each class includes 250 manually reviewed test images, rescaled to a maximum side length of 512 pixels. The presence of noise in some test images adds a significant challenge to the classification task.

\textbf{STL-10}~\citep{coates2011analysis} The STL-10 dataset is an image recognition dataset for developing unsupervised feature learning, deep learning, self-taught learning algorithms. It includes 10 classes: airplane, bird, car, cat, deer, dog, horse, monkey, ship, truck. Each images is $96 \times 96$ pixels, color. It contains 500 training images, 800 test images per class.

\textbf{DTD}~\citep{cimpoi14describing} DTD is a texture database, consisting of 5640 images, organized according to a list of 47 terms (categories) inspired from human perception. There are 120 images for each category. Image sizes range between $300 \times 300$ and $640 \times 640$, and the images contain at least 90\% of the surface representing the category attribute.

\textbf{Caltech-101}~\citep{fei2006one} Pictures of objects belonging to 101 categories. About 40 to 800 images per category. Most categories have about 50 images. 

It is worth noting that, due to our ultra-high computational efficiency, we can rapidly process entire test data in a short time. In contrast, for large-scale datasets (e.g., ImageNet-1k), Li's DC and Clark's DC only select part of the data for validation. The impressive results from Table~\ref{table:gdcbenchmark} underscore the robustness of GDC in achieving effective zero-shot classification under various challenges.

\subsection{Prompt Templates}
\label{sec:templates}
We use eight different prompt templates:'a photo of a \{\}', 'itap of a \{\}', 'a bad photo of the \{\}', 'a origami \{\}', 'a photo of the large \{\}', 'a \{\} in a video game','art of the \{\}', 'a photo of the small \{\}', replacing the placeholder '\{\}' with class name. We use each prompt template to generate 30 reference images averagely.

\section{Ablation study on regularization values}
\label{regularization}
We conduct ablation experiments of regularization value across various datasets, including ImageNet, CIFAR-10/100, Caltech-101, DTD and STL-10. The results are shown in Table~\ref{table:regularization}.
\begin{table}[h]
\begin{center}
\begin{small}
\resizebox{0.48\textwidth}{!}{
\begin{tabular}{c|ccc|c}
\toprule
 Regularization value & $1e^{-4}$ & $1e^{-6}$ & $1e^{-8}$ & $1e^{-10}$ \\
\midrule
Imagenet & 71.23 & 71.40 &  \textbf{71.44} & NaN\\
CIFAR10 & \textbf{97.92} & 96.96 &  96.80 & NaN \\
CIFAR100 & 82.95 & 83.92 & \textbf{84.02} &  NaN \\
Caltech-101 & 90.96 & 91.28 & \textbf{91.41} &  NaN \\
DTD & 48.03 & \textbf{48.24} & 48.19 &  NaN \\
STL-10 & 95.00 & \textbf{96.46} & 96.45 &  NaN \\
\bottomrule
\end{tabular}
}
\end{small}
\end{center}
\caption{The influence of regularization value, $\epsilon$, on classification accuracy. The lower regularization value correlates with the higher accuracy. If the value is lower than $1^{-10}$, inversing $\Sigma + \epsilon I $ is not stable.}
\label{table:regularization}
\end{table}


\section{Supplementary Discussion}

\textbf{Normality Test of DINOv2 Features} We assess the normality of DINOv2 feature representations for each class of generated images by performing Principal Component Analysis (PCA) followed by the Shapiro-Wilk normality test \citep{shapiro1965analysis}. We find that $60.01\%$  of the components are approximately Gaussian according to the statistical tests for ImageNet. 

Specifically, for each class, we first extract the DINOv2 feature representations \( e \) and compute their covariance matrix. Performing eigenvalue decomposition on this matrix yields the principal components, which are sorted by descending eigenvalues to prioritize the directions with the highest variance. We then project \( e \) onto these principal components. To evaluate normality along each component, we apply the Shapiro-Wilk test to each dimension in the projected space, yielding a p-value that reflects the likelihood that the distribution is Gaussian. A p-value above 0.05 suggests that normality cannot be rejected in that dimension. This approach allows us to analyze the normality characteristics across classes within the principal component space.

\textbf{Shapiro–Wilk test} To better understand the normality test, we present the calculation of the Shapiro–Wilk test. The Shapiro–Wilk test~\citep{shapiro1965analysis} tests the null hypothesis that a sample $x_1, \dots, x_n$ came from a normally distributed population. The test statistic is:
\begin{equation}
    W = \frac{\left(\sum_{i = 1}^na_ix_{(i)}\right)^2}{\sum_{i = 1}^n(x_i - \overline x)^2},
\end{equation}
$x_{(i)}$ with parentheses enclosing the subscript index $i$ is the $i_{th}$ order statistic, i.e., the $i_{th}$-smallest number in the sample. $\overline x$ is the sample mean. The coefficients $a_i$ are given by:
\begin{equation}
    (a_1,\dots, a_n) = \frac{m^TV^{-1}}{C},
\end{equation}
where $C$ is a vector norm: $||V^{-1}m|| = (m^TV^{-1}V^{-1}m)$, and the vector m $(m_1,\dots,m_n)^T$ is made of the expected values of the order statistics of independent and identically distributed random variables sampled from the standard normal distribution. The $V$ is the covariance matrix of those normal order statistics.

\textbf{Cholesky decomposition} We define the Cholesky decomposition~\citep{benoit1924note} of the precision matrix as the precision Cholesky term for the GMM. This decomposition represents a Hermitian, positive-definite matrix as the product of a lower triangular matrix and its conjugate transpose. It facilitates efficient numerical computations and is widely applied in solving systems of linear equations, linear least squares problems, nonlinear optimization, Monte Carlo simulations, matrix inversion, and more.

The Cholesky decomposition of a Hermitian positive-definite matrix $\mathbf{A}$ is expressed as $\mathbf{A} = \mathbf {LL}^*$, where $\mathbf{L}$ is a lower triangular matrix with real and positive diagonal entries, and $\mathbf{L^*}$ denotes the conjugate transpose of $\mathbf{L}$. This decomposition is unique for every Hermitian positive-definite matrix.  When A is a real symmetric positive-definite matrix, the Cholesky decomposition can be written $\mathbf {A} =\mathbf {LL} ^{\mathsf {T}}$, where $\mathbf{L}$ is a real lower triangular matrix with positive diagonal entries.

There are various approaches for calculating the Cholesky decomposition. We present two widely used approaches. The first one is the Cholesky algorithm, employed to compute the decomposition matrix $\mathbf{L}$, is an modified variant of Gaussian elimination.
The recursive algorithm starts with \( i := 1 \) and
\[
\mathbf{A}^{(1)} := \mathbf{A}.
\]
At step \( i \), the matrix \( \mathbf{A}^{(i)} \) has the following form:
\[
\mathbf{A}^{(i)} = 
\begin{pmatrix}
\mathbf{I}_{i-1} & 0 & 0 \\
0 & a_{i,i} & \mathbf{b}_i^* \\
0 & \mathbf{b}_i & \mathbf{B}^{(i)}
\end{pmatrix},
\]
where \( \mathbf{I}_{i-1} \) denotes the identity matrix of dimension \( i-1 \). If the matrix \( \mathbf{L}_i \) is defined by
\[
\mathbf{L}_i := 
\begin{pmatrix}
\mathbf{L}_{i-1} & 0 & 0 \\
0 & \sqrt{a_{i,i}} & 0 \\
0 & \frac{1}{\sqrt{a_{i,i}}} \mathbf{b}_i & \mathbf{I}_{n-i}
\end{pmatrix},
\]
then \( \mathbf{A}^{(i)} \) can be written as
\[
\mathbf{A}^{(i)} = \mathbf{L}_i \mathbf{A}^{(i+1)} \mathbf{L}_i^*,
\]
where
\[
\mathbf{A}^{(i+1)} =
\begin{pmatrix}
\mathbf{I}_{i-1} & 0 & 0 \\
0 & 1 & 0 \\
0 & 0 & \mathbf{B}^{(i)} - \frac{1}{a_{i,i}} \mathbf{b}_i \mathbf{b}_i^*
\end{pmatrix}.
\]
Note that \( \mathbf{b}_i \mathbf{b}_i^* \) is an outer product. This is repeated for \( i \) from 1 to \( n \). After \( n \) steps, \( \mathbf{A}^{(n+1)} = \mathbf{I} \) is obtained, and hence, the lower triangular matrix \( \mathbf{L} \) sought for is calculated as
\[
\mathbf{L} := \mathbf{L}_1 \mathbf{L}_2 \cdots \mathbf{L}_n.
\]

The second approaches is the Cholesky–Banachiewicz and Cholesky–Crout algorithms. If the equation
\[
\mathbf{A} = \mathbf{L} \mathbf{L}^T =
\begin{pmatrix}
L_{11} & 0 & 0 \\
L_{21} & L_{22} & 0 \\
L_{31} & L_{32} & L_{33}
\end{pmatrix}
\begin{pmatrix}
L_{11} & L_{21} & L_{31} \\
0 & L_{22} & L_{32} \\
0 & 0 & L_{33}
\end{pmatrix},
\]
\[
= 
\begin{pmatrix}
L_{11}^2 & L_{21}L_{11} & L_{31}L_{11} \\
L_{21}L_{11} & L_{21}^2 + L_{22}^2 & L_{31}L_{21} + L_{32}L_{22} \\
L_{31}L_{11} & L_{31}L_{21} + L_{32}L_{22} & L_{31}^2 + L_{32}^2 + L_{33}^2
\end{pmatrix},
\]
is written out, the following is obtained:
\[
\mathbf{L} =
\begin{pmatrix}
\sqrt{A_{11}} & 0 & 0 \\
A_{21} / L_{11} & \sqrt{A_{22} - L_{21}^2} & 0 \\
A_{31} / L_{11} & (A_{32} - L_{31}L_{21}) / L_{22} & \sqrt{A_{33} - L_{31}^2 - L_{32}^2}
\end{pmatrix}.
\]
And therefore the following formulas for the entries of \( \mathbf{L} \):
\[
L_{j,j} = (\pm) \sqrt{A_{j,j} - \sum_{k=1}^{j-1} L_{j,k}^2},
\]
\[
L_{i,j} = \frac{1}{L_{j,j}} \left( A_{i,j} - \sum_{k=1}^{j-1} L_{i,k} L_{j,k} \right), \quad \text{for } i > j.
\]

For complex and real matrices, inconsequential arbitrary sign changes of diagonal and associated off-diagonal elements are allowed. The expression under the square root is always positive if \( \mathbf{A} \) is real and positive-definite. So it now is possible to compute the \((i,j)\) entry if the entries to the left and above are known.

\textbf{More visual results} We presents more visual results of feature distribution of CIFAR-10/100 and ImageNet.
\begin{figure}[th]
    \centering
    \includegraphics[width=0.9\linewidth]{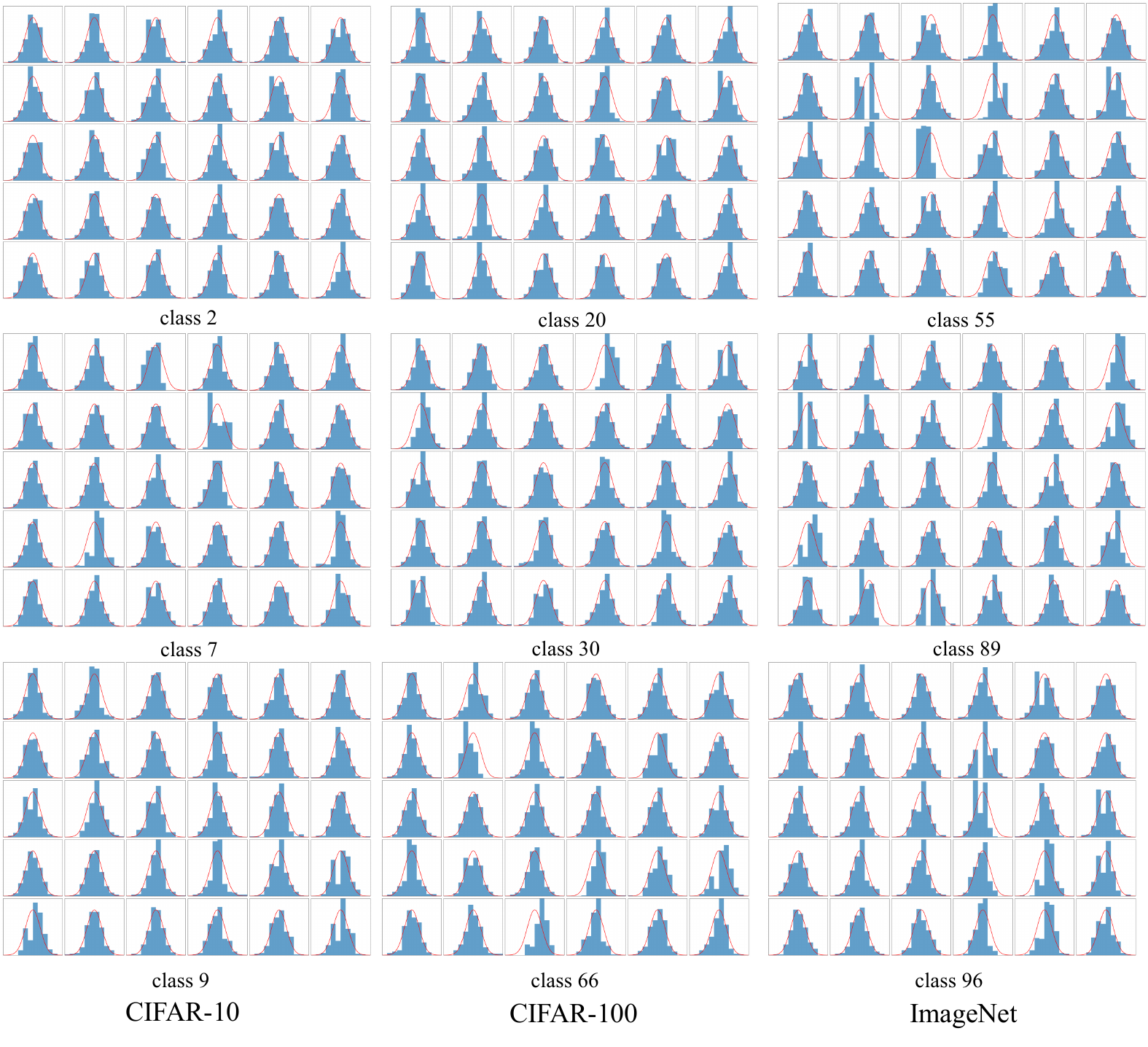}
    \caption{We visualize the distribution of image features for random 30 PCA components from three randomly picked classes on CIFAR-10/100 and ImageNet. }
    \label{fig:enter-label}
\end{figure}


\end{document}